\documentclass[journal]{IEEEtran}
\usepackage{moreverb,url}
\usepackage[colorlinks,bookmarksopen,bookmarksnumbered,citecolor=red,urlcolor=red]{hyperref}
\usepackage{amsmath}
\usepackage{amsbsy}
\usepackage{amssymb}
\usepackage{amsthm}
\usepackage{array}
\usepackage{verbatim}
\usepackage{subfigure}
\usepackage{color}
\usepackage{xcolor}
\usepackage{hyperref}
\hypersetup{urlcolor=black} 
\usepackage{mathtools}
\usepackage{theorem}
\usepackage{pifont}
\usepackage{graphicx,float} 
\usepackage{tabularx,ragged2e,booktabs,caption}
\usepackage[linesnumbered,ruled,vlined]{algorithm2e}
\usepackage{multirow}
\usepackage{makecell}
\DeclareMathOperator*{\argmin}{arg\,min}

\begin{document}
\title{Fast and Accurate Data-Driven Simulation Framework for Contact-Intensive Tight-Tolerance Robotic Assembly Tasks}

\author{Jaemin Yoon, Minji Lee, Dongwon Son, and Dongjun Lee
\thanks{This research was supported by the Industrial Strategic Technology Development Program (20001045) of the Ministry of Trade, Industry $\&$ Energy (MOTIE) of Korea.}
\thanks{J. Yoon was with Seoul National University and is now with the Robot Center, Samsung Research, Seoul, Republic of Korea (e-mail: jae$\_$min.yoon@samsung.com).}
\thanks{M. Lee, and D. J. Lee are with the Department of Mechanical Engineering, IAMD (Institute of Advanced Machines and Design) and IER (Institute of Engineering Research), Seoul National University, Seoul, Republic of Korea (e-mail: mingg8@snu.ac.kr; djlee@snu.ac.kr). Corresponding author: Dongjun Lee.}
\thanks{D. Son was with Seoul National University and is now with AI Method Team, Samsung Research, Seoul, Republic of Korea (e-mail: dongwon.son@samsung.com).}
}

\maketitle

\begin{abstract}
We propose a novel fast and accurate simulation framework for contact-intensive tight-tolerance robotic assembly tasks.  
The key components of our framework are as follows: 
1) data-driven contact point clustering with a certain variable-input network, which is explicitly trained for simulation accuracy (with real experimental data) and able to accommodate complex/non-convex object shapes;
2) contact force solving, which precisely/robustly enforces physics of contact (i.e., no penetration, Coulomb friction, maximum energy dissipation) with contact mechanics of contact nodes augmented with that of their object;
3) contact detection with a neural network, which is parallelized for each contact point, thus, can be computed very quickly even for complex shape objects with no exhaust pair-wise test; and
4) time integration with PMI (passive mid-point integration \cite{KimIJRR17}), whose discrete-time passivity improves overall simulation accuracy, stability, and speed. 
We then implement our proposed framework for two widely-encountered/benchmarked contact-intensive tight-tolerance tasks, namely, peg-in-hole assembly and bolt-nut assembly, and validate its speed and accuracy against real experimental data. It is worthwhile to mention that our proposed simulation framework is applicable to other general contact-intensive tight-tolerance robotic assembly tasks as well. We also compare its performance with other physics engines and manifest its robustness via haptic rendering of virtual bolting task.
\end{abstract}

\begin{IEEEkeywords}
Contact clustering, bolting, data-driven approach, multi-point contacts, peg-in-hole, robotic assembly simulation, tight-tolerance.
\end{IEEEkeywords}

\IEEEpeerreviewmaketitle

\section{Introduction}.
\label{sec:introduction}
\IEEEPARstart{W}{ith} the recent advances of robot hardware and software technologies, many attempts are being made to bring the robots out from research labs to real industrial settings (e.g., \cite{WangIJAC18, SanchezIJRR18}). 
Among those industrial applications, contact-intensive and tight-tolerance robotic assembly tasks (e.g., peg-in-hole assembly \cite{ParkTIE17}, bolt-nut assembly \cite{XuAccess19}, snap connector assembly \cite{MJ_IROS21}) are ubiquitous, but deemed fairly challenging, since they typically require complex simultaneous motion and force control to cope with geometric and material uncertainties (e.g., in pose, shape, size, friction, etc.) while preventing jamming, slippage, or even breakage of parts.

To develop a complex control strategy for the contact-intensive tight-tolerance tasks, which is also robust (i.e., generalizable) against those uncertainties as well, in a safe and cost-effective manner, a fast and accurate simulator is desirable. 
This is particularly so for the recently-burgeoning technique of reinforcement learning (RL, e.g., \cite{SonIROS20, levine2020offline}), which is well-known for the ability to generate complex and robust control policies even for mathematically-vague/intractable tasks, yet, also well-known for the necessity of excessively-large datasets (i.e., data inefficiency) for its learning (e.g., slightly less than a million dataset for (simpler) picking task \cite{LevineIJRR18}).
Such fast and accurate simulators will also greatly facilitate the development process of the classical model-based motion/force controls as well.

\begin{figure}
	\centering
	\includegraphics[width=0.485\textwidth]{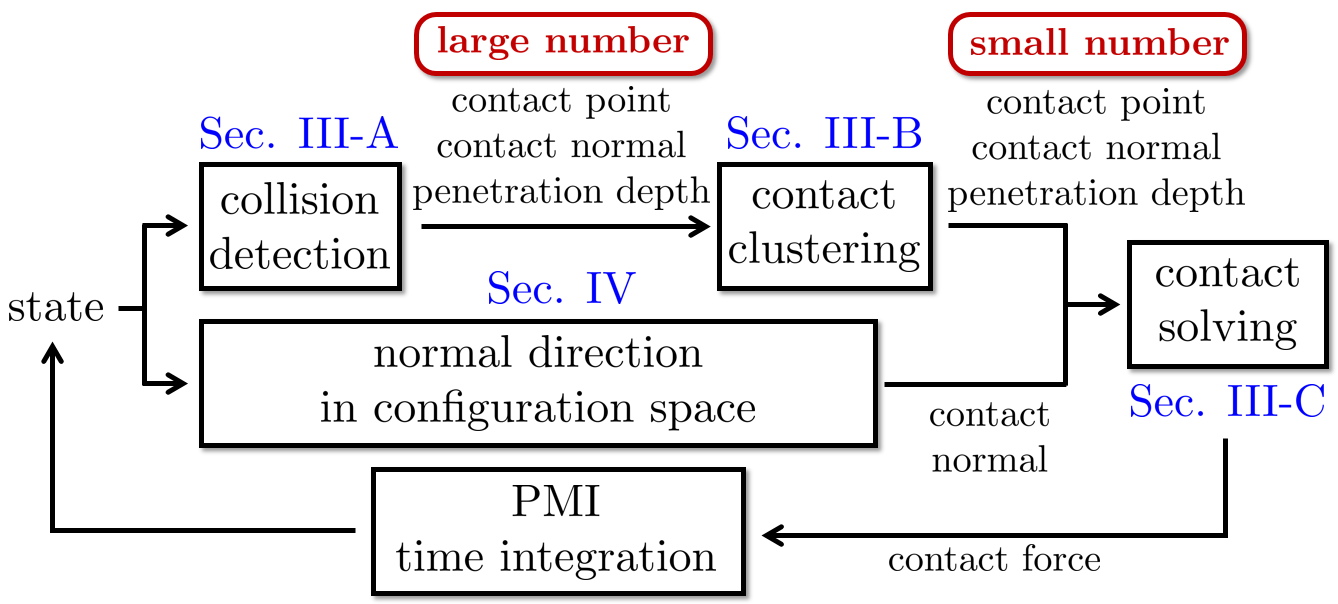}
	\caption{Pipelines of the contact simulation framework.}
	\label{fig:diagram}
    \vspace{-0.15in}
\end{figure}

In this paper, we propose a novel framework for the complete pipelines for fast (i.e., near real-time) and accurate (i.e., matching well with real experimental data) simulation of contact-intensive tight-tolerance assembly tasks - see Fig. \ref{fig:diagram}. 
We particularly focus on the peg-in-hole task and the bolt-tightening task, arguably the two most widely-encountered/benchmarked assembly tasks, with the framework proposed for them applicable to other general contact-intensive tight-tolerance robotic assembly tasks as well.  
These two tasks are challenging, since, due to the tight-tolerance and complex/non-convex contact geometry, the contact behavior is very sensitive/precarious to the geometric uncertainty and any successful control should in essence be able to accurately predict that.  
Simulating the two tasks fast and accurately is even more challenging, since, from its discrete implementation, the simulation can not only severely chatter due to the tight-tolerance and complex geometry as stated above, but also lose relevance with the real physics (e.g., penetration) or even completely go unstable. 
To our knowledge, this fast and accurate simulation of the two tasks is achieved in this paper for the first time with typical open-source simulators (e.g., Vortex \cite{Vortex}, ODE \cite{SmithODE05}, Bullet \cite{CoumansBULLET13}, MuJoCo \cite{TodorovMUJOCO12}) not able to do so (see Fig. \ref{fig:snapshot_opensource}) likely due to their aiming for general simulation scenarios rather than being optimized for the contact-intensive/tight-tolerance tasks unlike our simulation framework.

The key innovations of our framework to achieve this, which are applicable to other contact-intensive tight-tolerance robotic assembly tasks as well, are as follows:
\begin{itemize}
\item {\bf Data-driven contact clustering with IN (interaction network)}: The process of contact clustering (i.e., reduce a large number of contact points into a manageable number of clusters with their contact normals) is necessary to attain practically fast-enough simulation speed for the contact-intensive tight-tolerance tasks. Less recognized fact yet is that this clustering also greatly affects the contact simulation accuracy, particularly when the contact geometry is complex and the contact points are many. 
This contact clustering, however, has been tacked with not much heed on the simulation accuracy in rather an ad-hoc fashion (e.g., use first sixteen contact points in MuJoCo \cite{TodorovMUJOCO12}, select four contact points to maximize their covered area in Bullet \cite{CoumansBULLET13}) or in simply Euclidean-geometric fashion 
(e.g., $K$-means clustering \cite{JainCSUR99}). 
In contrast, we formulate this contact clustering as a data-driven machine learning process trained to match the experimental data given the dynamic information (i.e., pose, twist, and wrench) of all the contact points, thereby, significantly improving the contact simulation accuracy. 
We also adopt IN architecture \cite{BattagliaANIPS16} for this to accommodate 
vastly-varying contact point set for the two tasks due to their complex/non-convex geometries, that cannot be addressed with other typical learning architectures (e.g., MLP (Multi-Layer Perceptron \cite{KimICRA19}));
\begin{figure}
	\centering
	\includegraphics[width=0.45\textwidth]{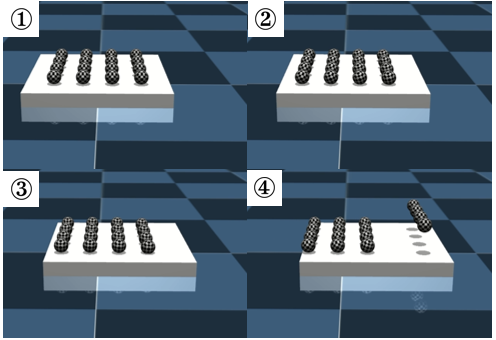}
	\caption{Snapshots of simulation of multiple balls on a sideway-moving slab: the balls suddenly jump out from the flat slab due to approximation of penetration-related constraints.}
	\label{fig:mujoco_bouncing}
    \vspace{-0.15in}
\end{figure}

\item {\bf Constraint/energy-based robust contact solving}: Given the clustered points (and normals), the process of contact solving computes the contact force for each point using relevant physics principles. For this, two main methodologies exist: penalty-based method (e.g., \cite{DrumTVCG08, otaduy2006modular}) and constraint-based method (e.g., \cite{lloyd2005fast, HorakRAL19}), with the latter preferred when the accuracy is concerned, since the former, from its relying on virtual springs, typically cannot precisely enforce physical principles of contact (e.g., Signorini condition for no penetration \cite{StewartSIAM20}, Coulomb friction cone constraint) with the contact force also greatly varying per user-tuned parameters of the springs even for the same motion. 
Thus, many works adopt the constraint-based method, yet, relaxing some of the principles for computation speed/efficiency (e.g., Signorini condition in MuJoCo \cite{TodorovMUJOCO12}, friction cone constraint in Vortex \cite{Vortex}), thus, can exhibit inaccurate and even unstable contact simulation (see Fig. \ref{fig:mujoco_bouncing}). 
In contrast, we strictly enforce those principles of contact, and, for this, our utilization of PMI (passive mid-point integration \cite{KimIJRR17}) is instrumental,
since it provides us ``rooms'' for precisely solving the contact principles thanks to its superior speed and stability. 
This PMI also allows us to duplicate salient energetics of contact (i.e., maximum energy dissipation principle \cite{PreclikCM18}) in the discrete simulation, further enhancing the accuracy of contact simulation.  
We also augment this node-space (N-space) contact solving (for each contact point (node)) with the configuration-space (C-space) contact solving (for the object configuration), which turns out to significantly improve the robustness of penetration prevention (see Fig. \ref{fig:Ceffect2}) and is another novelty of our proposed framework. 

\item {\bf Parallelized contact detection with NN (neural network)}: The process of contact detection determines whether or not objects intersect with each other and finds the set of their contact points, normals, and penetration depths. 
Among the simulation pipelines in Fig. \ref{fig:diagram}, this contact detection is perhaps most standardized with some standard approaches (e.g., \cite{GilbertJRA88, SnethenGPG08}) embedded in most open-source simulators.  
These standard approaches, however, are typically too slow for our target tasks (to be real-time), since, due to the complex/non-convex object shapes with tight-tolerance (e.g., bolt/nut), a large number of meshes/nodes are necessary for accurate contact simulation, yet, those standard approaches require exhaustive pair-wise test between those meshes.  
In contrast, we extend the idea of C-space data-driven contact detection method of \cite{SonIROS20} to the N-space of our target tasks, which is free from such exhaust pair-wise test and rather exploits NN to model the penetration depth and normal for each contact node.
We further parallelize the calculation of NNs for each contact node utilizing a graphic processing unit (GPU), thereby, significantly accelerating contact detection process (e.g., about ten times faster than the state-of-the-art FCL (flexible collision library \cite{PanICRA12}) - see Table 1). 
It is also noteworthy that the training of this NN can be done completely off-line only using CAD models of the involved objects, since the contact detection is purely a geometric process depending only on the relative pose between the objects. 

\item {\bf Time integration with PMI \cite{KimIJRR17}}: This process integrates the discrete-time rigid object dynamics with the contact force as solved above. 
For this, various types of integrators have been used: explicit Euler integrator (EEI, \cite{atkinsonJWS08}), implicit Euler integrator (IEI, \cite{butcherWOL08}), and semi-implicit Euler integrator (SEI, \cite{hairerAN03}).
These integrators, however, do not guarantee simulation stability (e.g., EEI, SEI) or simulation speed (e.g., IEI), and all of them do not enforce passivity (or energy conservation), thus, exhibiting proneness to simulation instability or dissipativity. 
In contrast, we adopt our recently-proposed PMI \cite{KimIJRR17} for the time integration pipeline in Fig. \ref{fig:diagram}, which, due to its discrete-time passivity, can maintain simulation stability even against very light/stiff dynamics or variable time-step, frequently-arising in tight-tolerance contact simulation, while also retaining the simulation speed with no time-consuming nonlinear function solving (from its being non-implicit) and allowing us to apply the concept of energetics in the discrete-time domain (e.g., contact solving with maximum energy dissipation principle in Sec. 3.3), thereby, improving the overall accuracy and speed of our proposed contact simulation framework. 
\end{itemize}

Along the same vein as this paper, an accuracy-optimized data-driven contact clustering for rigid-body contact simulation was proposed in \cite{KimICRA19}.
The contact clustering of \cite{KimICRA19} however is applicable only for simple primitive-primitive contact scenarios (e.g., cylinder-on-plane contact in \cite{KimICRA19}), since, from its being based on the (static) architecture of MLP \cite{haykin2010neural} with all the (pre-defined) contact nodes of all the (pre-defined) meshes as its input, 
for the target bolting and peg-in-hole tasks with complex/non-convex shape objects, 
it is impossible to learn and real-time compute its MLP, which will become 
exceedingly large with those massive number of nodes carried with all the time. 
On the other hand, our proposed data-driven contact clustering, from its instead being based on the (variable) architecture of IN, is scalable for such objects for our target tasks with very complex/non-convex geometries.  
The ideas of maximum energy dissipation contact force generation, combining C-space with N-space to robustify penetration prevention, and parallelized contact detection with NN are also presented in this paper for the first time.

It is also worthwhile to mention that our proposed simulation framework is generalizable at least to some extent (e.g., across different bolts/nuts in Sec. 5.5).
This then implies that, if one collects experimental data and trains the contact clustering networks for only one assembled part, 
it can be used for all the other parts with no further (typically expensive) experiments/training as long as the parts to assemble are with controlled tolerance and bounded uncertainties.
This would be particularly suitable for mass production line (e.g., assembly/bolting task in car factory with jig-fixture/machining tolerance).
Recall that all the other pipelines of our framework in Fig. \ref{fig:diagram} do not require experiments with the data-driven contact detection necessitating off-line learning only with the CAD data of the objects.  
To show the speed, robustness, and versatility of our proposed simulation framework, we also apply it to the real-time haptic rendering of bolting task with arbitrary human command (Sec. 5.6), which, to our knowledge, is also achieved in this paper for the first time. We also make our simulation framework (closed-source) open to public: \url{https://github.com/INRoL/inrol_sim_peginhole}, \url{https://github.com/INRoL/inrol_sim_bolting}

The rest of the paper is organized as follows. Sec. 2 presents some preliminary materials for the contact simulation. Sec. 3 presents the key components of our proposed simulation framework: parallelized contact detection with NN, accuracy-optimized contact clustering with IN, and constraint/energy-based contact solving.
Sec. 4 then explains how to augment the N-space contact simulation in Sec. 3 with the C-space contact formalism to improve the robustness of the overall contact simulation.
Sec. 5 presents validation, demonstration, and haptic-application of our proposed simulation framework against real experimental results and compares it with other physics engines. Summary and some comments on the future research directions are then given in Sec. 6.

\section{Preliminary}
\label{sec:pre}
Typically, the geometric primitives of the object are modeled by triangle mesh elements in three-dimensional space for contact detection in the contact simulation. This mesh element consists of the nodes (or points), which are the vertexes of the mesh. The friction force required to generate the contact force is straightforward to formulate for these three-dimensional nodes. Thus, we consider the contact model in this three-dimensional space consisting of the nodes, which we call node space (N-space) from now on. For the N-space, we define $x_i \in \Re^3$ to be the position of each node.

To generate the contact force in this N-space, we first consider the discrete-time dynamics of their single rigid object with multi-point contact by applying passive midpoint integration (PMI \cite{KimIJRR17}), which can be written as
\begin{align}
\label{eq:dynamics}
M \frac{V_{k+1} - V_k}{T_k} + C_k(\omega_k) \frac{V_{k+1}+V_k}{2} = \sum_{i=1}^{N_c} J_{i,k}^T f_{i,k}
\end{align}
where $\star_k = \star(t_k)$, $T_k > 0$ is the update time step, $V_k := [v_k;w_k] \in \Re^6$ is the linear and angular velocity, $M := \text{diag}[m I_{3\times 3}, J] \in \Re^{6 \times 6}$ is the inertia matrix, $m > 0$, $J \in \Re^{3 \times 3}$ are the mass and moment of inertia, $C_k := \text{diag}[0_{3 \times 3}, -S(J \omega_k)] \in \Re^{6 \times 6}$ is the Coriolis matrix, $S(\cdot)$ is the skew-symmetric operator with $S(a) b = a \times b, a, b \in \Re^3$, and $N_c$ is the number of contact nodes with $J_{i,k} \in \Re^{3 \times 6}$ and $f_{i,k} \in \Re^3$ being respectively the contact Jacobian and the contact force acting on the $i$-th contact node $x_i$.  The contact force $f_{i,k}$ is defined following the point contact model s.t.,
\begin{align*}
f_{i,k} = \vec{n}_{i,k} \lambda^n_{i,k} + \vec{t}_{i,k} \lambda^t_{i,k}
\end{align*}
where $\lambda^n_{i,k} \geq 0$, $\lambda^t_{i,k} \in \Re^2$ are the normal and tangential contact force, and $\vec{n}_{i,k} \in \Re^3, \vec{t}_{i,k} \in \Re^{3 \times 2}$ are the normal and tangential direction vectors of the contact surface at $x_i$.

The contact force can be calculated by considering rigid body dynamics \eqref{eq:dynamics} with the following physical principles: 1) Signorini condition \cite{StewartSIAM20} that the contact force for each node is generated in the direction of eliminating collision, i.e.,
\begin{align}
\label{eq:Signorini}
0 \leq \lambda^n_{i,k} \perp v^n_{i,k+1} \geq 0
\end{align}
where $v^n_{i,k+1} := \vec{n}_{i,k}^T J_{i,k} V_{k+1} \in \Re$ is the velocity along the normal direction of $x_i$ at next time step, when we use the velocity-level constraint for plastic contact, and the operator $\perp$ means that only one of $\lambda^n_{i,k}$ and $v^n_{i,k+1}$ is non-zero, which prevents contact force from being generated when objects move away from each other; 2) Coulomb friction cone constraint s.t.,
\begin{align}
\label{eq:Coulomb}
\Vert \lambda^t_{i,k} \Vert_2 \leq \mu \lvert \lambda^n_{i,k} \rvert
\end{align}
where $\mu$ is the friction coefficient; 3) the maximum energy dissipation principle \cite{PreclikCM18} that the friction force is generated in the direction, which maximizes the energy dissipation of the object.   The details on the calculation of contact force are explained in Sec. 3.3.

As the number of contact points increases due to complicated object geometry, simulation involves a high computational load because the number of constraints to be considered when calculating the contact force increases. Also, simulation can be numerically unstable because the contact problem can be overdetermined as the number of contact constraints to be satisfied increases. For real-time simulation, contact clustering, which determines a small number of clustered contact nodes and their normals, is necessary. Also, this clustering affects not only the simulation speed, but also the accuracy of the contact force. The analytical formulation of how contact clustering affects the accuracy of the contact force, however, is impossible in most cases. To ensure the accuracy of the contact force during contact clustering, we develop a data-driven contact clustering framework to be explained in Sec. 3.2.
For this, let us here define the original contact point (node) and normal $(x^o_i,n^o_i)$ for $i=\{1,\cdots,N_c\}$, where $N_c$ is the number of original contact nodes, and those of their clustered contact nodes $(x^r_j,n^r_j)$ for $j=\{1,\cdots,M_c\}$, where $M_c < N_c$ is the number of clustered contact nodes.

Even with the data-driven contact clustering in Sec. 3.2 with the accuracy of the contact force explicitly considered, penetration can still occur in contact nodes other than the clustered contact nodes, since only the contact constraints (e.g., prevent penetration with \eqref{eq:Signorini}) for the clustered contact nodes will be activated. To increase the robustness against this problem, we develop the contact model in SE(3) configuration space (C-space) of the object, which essentially allows us to incorporate collision information of all parts of the object. For this C-space representation, we define the configuration $q := [x_c;\phi] \in \Re^6$, where $x_c \in \Re^3$ is the center position of the object (e.g., peg/nut center) and $\phi \in \Re^3$ is the XYZ-Euler angles. The details on C-space contact model will be explained in Sec. 4.

\section{Node Space Contact Simulation}
\label{sec:Nspace}
\begin{figure}
	\centering
	\includegraphics[width=0.485\textwidth]{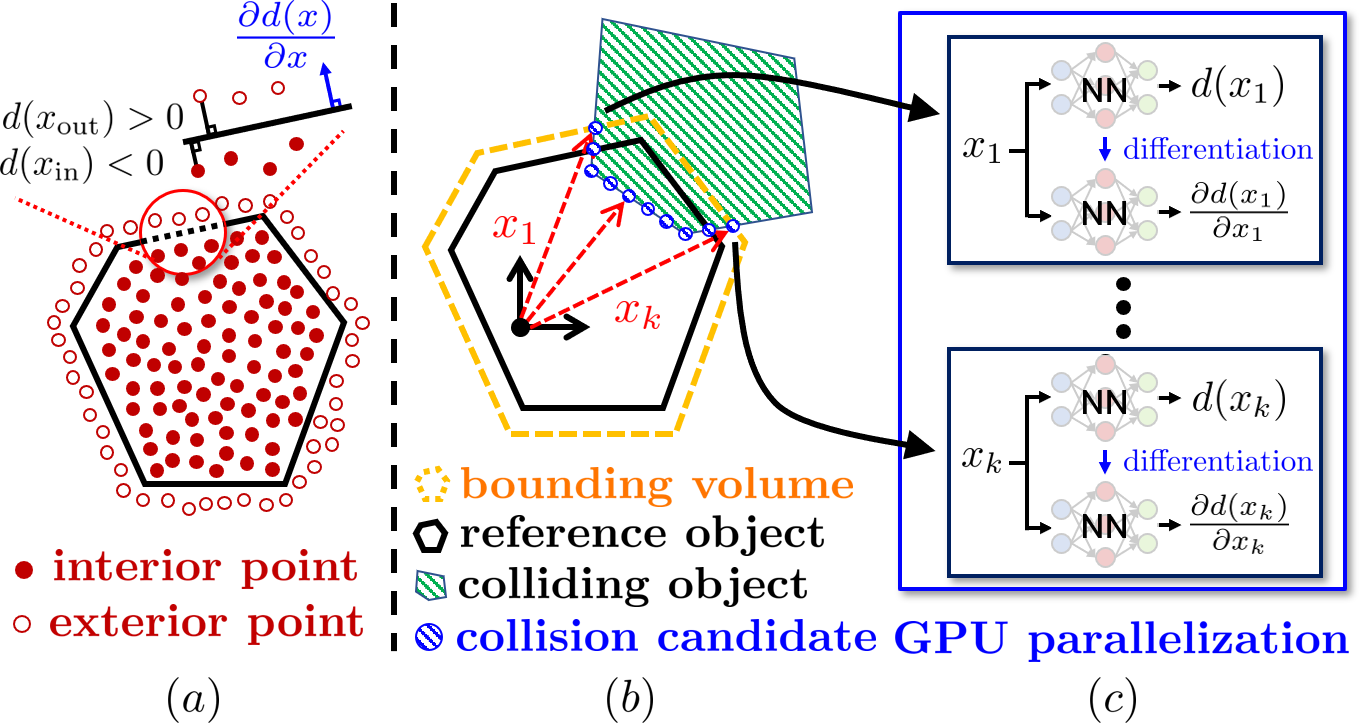}
	\caption{Illustration of contact detection process with NN: (a) data collection to model the penetration depth; (b) geometry when contact between two objects occurred; (c) parallelized NN computation for each contact node $x_i$.}
	\label{fig:CD}
    \vspace{-0.15in}
\end{figure}

\subsection{Contact Detection using Neural Network}
\label{subsec:CD_Nspace}
To reduce the simulation computing time, we first develop the learning-based contact detection method that can be parallelized for each (original) contact node. More precisely, we model the penetration depth of each contact node $x_i$ as $d_i(x_i) \in \Re$ w.r.t the reference object with a neural network (NN). Here, the reason why we adopt NN to model the penetration depth is that we need a general function approximation technique applicable to various objects regardless of their shapes, convex or concave. 
This NN modeling of penetration depth is done by generating many interior and exterior points, computing their $d_i(x_i)$ and then training the NN with a standard regression problem for each $x_i$ as shown in Fig. \ref{fig:CD} (a). 
Note that this process can be done off-line and purely geometrically using only the CAD models of the involved rigid objects, since the contact detection only depends upon the relative pose between the objects.

Once we construct $d_i(x_i)$, we also calculate the contact normal $\partial d_i(x_i)/\partial x_i \in \Re^3$ by finding the direction, in which the contact disappears as $d_i(x_i)$ increases. Using this trained NN can reduce the computational load as compared to performing an exhaustive pair-wise test of standard contact detection methods (e.g., \cite{GilbertJRA88, SnethenGPG08}), but it can still be computationally heavy since the penetration depth and contact normal should be calculated for all vertexes of the object. Thus, we further accelerate this process by applying broad phase contact detection (i.e., calculate $d_i(x_i)$ and $\partial d_i(x_i)/\partial x_i$ only for the collision candidate points inside the bounding volumes) and GPU parallelization for each contact node as illustrated in Fig. \ref{fig:CD} (b) and (c). By using this proposed data-driven/parallelized approach, we can significantly accelerate the contact detection speed (e.g., about ten times faster than FCL \cite{PanICRA12} - see Table 1).

\subsection{Data-Driven Contact Clustering}
\label{subsec:CC_Nspace}
\begin{figure}
	\centering
	\includegraphics[width=0.485\textwidth]{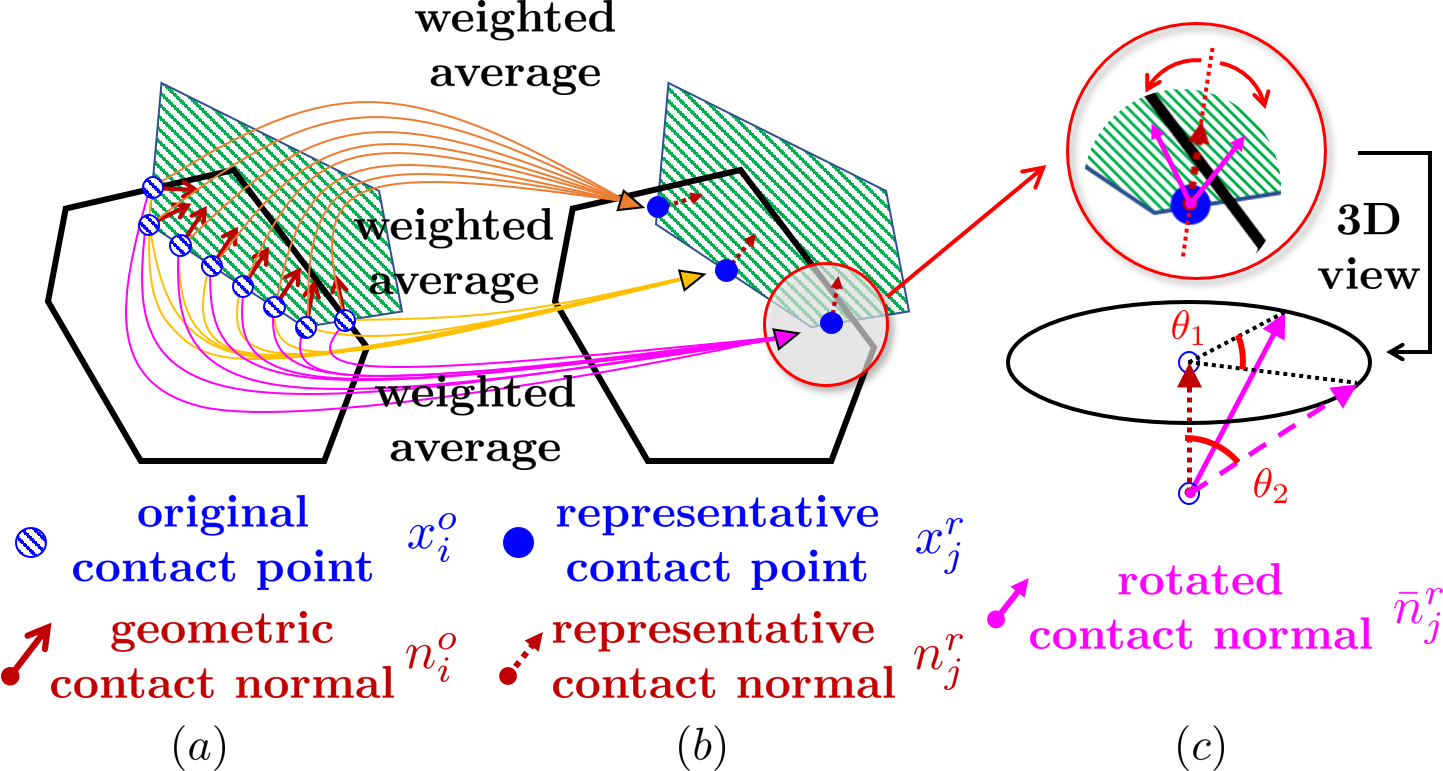}
	\caption{Illustration of data-driven contact clustering: (a,b) the process determines the clustered (representative) contact nodes; (c) their rotated contact normals.}
	\label{fig:WA}
    \vspace{-0.15in}
\end{figure}

To achieve real-time simulation while enhancing the accuracy of the contact force, we propose accuracy-optimized data-driven contact clustering. To determine the clustered contact nodes $\mathcal{X}^r := \{x^r_1,\cdots,x^r_{M_c}\}$ and normals $\mathcal{N}^r := \{n^r_1,\cdots,n^r_{M_c}\}$, 
we compute the weighted average of the original contact nodes $\mathcal{X}^o := \{x^o_1,\cdots,x^o_{N_c}\}$ and normals $\mathcal{N}^o := \{n^o_1,\cdots,n^o_{N_c}\}$. This process is depicted in Fig. \ref{fig:WA} (a) and (b).

More precisely, define the weight of $x^o_i$ for $x^r_j$ as $W^j_i$. To determine the weight, we design a NN with a set of weights of $x^o_i$ as output s.t.,
\begin{align*}
Y^W_i := \left[ W^1_i;\cdots;W^{M_c}_i \right] \in \Re^{M_c}
\end{align*}
The relation between $x^o_i, x^r_j$, and $Y^W_i$ by using the weighted average scheme is then given as follows:
\begin{align*}
\begin{bmatrix}
(x^r_1)^T\\
\vdots\\
(x^r_{M_c})^T
\end{bmatrix}=
\underbrace{
\begin{bmatrix}
\vert &  & \vert\\
Y^W_1 & \cdots & Y^W_{N_c}\\
\vert &  & \vert
\end{bmatrix}}_{=:\bar{Y}^W \in \Re^{M_c \times N_c}}
\begin{bmatrix}
(x^o_1)^T\\
\vdots\\
(x^o_{N_c})^T
\end{bmatrix}
\end{align*}
Here, if we use a simple weighted averaging, $\mathcal{X}^r$ may not exist on the surface of the object, where the actual contact nodes should exist. To address this, we calculate the weighted averaging on surfaces \cite{rustamovCGF09}, which use geodesic distance to determine the point on the contact surface, at which the weights of each contact node are considered.

Since it is intuitively reasonable that $\mathcal{X}^r$ is affected not only by geometric characteristics but also by dynamic characteristics, which can affect contact force, we choose the input $X^o_i$ for this NN as the dynamics-related variables of $x^o_i$ including their external force and moment, i.e.,
\begin{align}
\label{eq:input}
X^o_i := [x^o_i;q^o_i;v^o_i;w^o_i;f^o_i;\tau^o_i] \in \Re^{19}
\end{align}
where $q^o_i \in \Re^4, v^o_i \in \Re^3, w^o_i \in \Re^3$ are the quaternion, and linear/angular velocity of $x^o_i$, and $f^o_i \in \Re^3, \tau^o_i \in \Re^3$ are the external force/moment acting on $x^o_i$.

\begin{figure}
	\centering
	\includegraphics[width=0.485\textwidth]{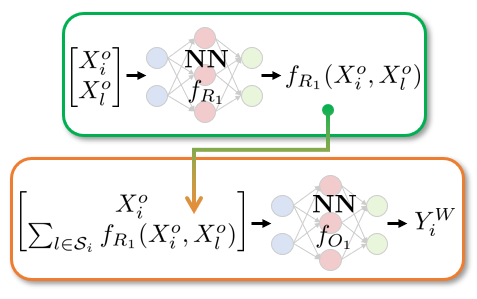}
	\caption{The structure of a neural network for contact clustering utilizing the structure of the interaction network.}
	\label{fig:IN}
    \vspace{-0.15in}
\end{figure}

We can then calculate $Y^W_i$ utilizing the structure of the interaction network (IN \cite{BattagliaANIPS16}):
\begin{align}
\label{eq:output}
Y^W_i &= f_{O_1}\Big(X^o_i, \sum_{l \in \mathcal{S}_i} f_{R_1} (X^o_i, X^o_l)\Big)
\end{align}
where $f_{R_1}$ is a NN, which predicts the effect of physical interaction between $x^o_i$ and $x^o_l$, $f_{O_1}$ is a NN, which predicts the effect of the dynamic characteristics of $x^o_i$, and $\mathcal{S}_i$ is the set containing only the contact nodes around $x^o_i$ with a size determined by the user. The NN structure for the contact clustering are shown in Fig. \ref{fig:IN}. The reason why we consider only the surrounding contact nodes without considering all the contact nodes is to limit the computational burden that increases as the number of original contact nodes increases. Here, we choose this IN, since the structure of IN can be applied regardless of the number of original contact nodes $N_c$ because $\sum_{l\in \mathcal{S}_i} f_{R_1} (X^o_i,X^o_l)$ in \eqref{eq:output} can be calculated regardless of $N_c$. It cannot be achieved by using MLP \cite{KimICRA19} or Long Short-Term Memory models (LSTM \cite{sak2014long}). Further, by using the IN, we can capture the effect of physical interaction between the dynamic characteristics of contact nodes, which can affect the contact force. 

Similar to the clustered contact nodes $\mathcal{X}^r$, the clustered contact normals $\mathcal{N}^r$ can also be determined through the weighted average scheme using the learned weights $\bar{Y}^W$. However, unlike the contact nodes, the original contact normals $\mathcal{N}^o$ are not properly defined due to the unmodeled geometry such as roughness of the contact surface, so unreliable $\mathcal{N}^r$ can be obtained even using the learned weights $\bar{Y}^W$. To resolve this, we design an additional NN that rotates $n^r_j$ considering the dynamics-related variables of $x^r_j$ as done above when determining $\mathcal{X}^r$ and the information on the clustered contact normal $\mathcal{N}^r$. For this, we set the output of this NN as the rotation angle s.t.,
\begin{align*}
Y^N_j := [\theta_j^1;\theta_j^2] \in \Re^2
\end{align*}
where $\theta_j^1,\theta_j^2$ are the angles to rotate $n^r_j$. Then the rotated contact normal $\bar{n}^r_j$ is determined by 
\begin{align*}
\bar{n}^r_j = R_n(\theta_j^1) R_t(\theta_j^2) n^r_j
\end{align*}
where $R_n(\theta_j^1)$ is the rotation matrix that rotates by $\theta_j^1$ with $n^r_j$ as the axis, and $R_t(\theta_j^2)$ is the rotation matrix that rotates by $\theta_j^2$ with the tangential vector perpendicular to $n^r_j$ as the axis. The rotation process is illustrated in Fig. \ref{fig:WA} (c). We then choose the input $\bar{X}^r_j := \left[X^r_j;n^r_j \right] \in \Re^{22}$ for this NN, where $X^r_j$ is the dynamics-related variables of $x^r_j$ similar to \eqref{eq:input}. 
We design the additional NN similar to \eqref{eq:output}, which is given as
\begin{align}
\label{eq:output2}
Y^N_j &= f_{O_2}\Big(\bar{X}^r_j, \sum_{l \neq j} f_{R_2} (\bar{X}^r_j, \bar{X}^r_l)\Big)
\end{align}
where $f_{R_2}$ is a NN, which predicts the effect of physical interaction between $x^r_j$ and $x^r_l$, and $f_{O_2}$ is a NN, which predicts the effect of the dynamic characteristics of $x^r_j$. In this NN, unlike the first NN \eqref{eq:output}, the information of all the clustered contact nodes is used instead of only the relationship with some inputs (i.e., $l \neq j$ in \eqref{eq:output2} instead of $l \in \mathcal{S}_i$ in \eqref{eq:output}). The reason is
that the number of inputs is limited because this NN calculates the output using only the clustered contact nodes, so there is no increase in the amount of computation as the number of original contact nodes from the contact detection increases.

When we use the proposed contact clustering, fast and accurate contact simulation is possible if NNs are well trained with real experimental data. However, the experimental contact points/normals needed for training the NNs cannot be directly measured due to the lack of appropriate sensors. Thus, we train the clustering NNs using the contact wrench measured at the center of the object, which can be directly measured using force/torque sensor. 
For this, we calculate the cost related to the error between the simulated and measured contact wrenches at the object center to train the NNs. Here, the simulated contact wrench is obtained by running the contact detection method in Sec. 3.1, contact clustering in Sec. 3.2, and contact solver in Sec. 3.3 all together. 
For training the NNs, we also utilize covariance matrix adaptation evolution strategy (CMA-ES \cite{KernNC04}), a type of gradient-free optimization technique, since the multi-point contact is in general not analytically differentiable.

\subsection{Contact Force Solving}
\label{subsec:CFG_Nspace} 
Maximum energy dissipation principle is well defined in the continuous-time domain with Coulomb friction law, that is,
the tangential friction force should always occur in the opposite direction to the tangential velocity.
It is however not the case of the discrete-time domain, since it is unclear what velocity at which time index is to use to determine the direction of the friction force. 
To duplicate this maximum energy dissipation principle precisely in the discrete-time domain, in this paper, 
we adopt the framework of PMI (passive mid-point integration  \cite{KimIJRR17}), which is discrete-time passive, thus, allow us to imbue our discrete-time simulation with the concept of energy conservation.

More precisely, consider the PMI formulation of the rigid-body dynamics  \eqref{eq:dynamics}. 
We can then write the discrete-time passivity (or energy conservation) relation s.t.,
\begin{align*}
\Big( \sum_{i=1}^{N_c} J_{i,k}^T f_{i,k} \Big)^T \Big( \frac{V_{k+1}+V_k}{2}\Big) T_k = E_{k+1} - E_k
\end{align*}
where $E_k := \frac{1}{2} V_k^T M V_k$ is the kinetic energy of the object. Then, the maximum energy dissipation principle for each contact can be achieved by minimizing the energy produced by only the tangential friction force, which can be formulated as the following optimization problem:
\begin{align}
\label{eq:opt}
\lambda^t_{i,k,\text{opt}} = \argmin_{\lambda^t_{i,k}} ~(J_{i,k}^T \vec{t}_{i,k} \lambda^{t}_{i,k} )^T \Big(\frac{V_{k+1}+V_k}{2}\Big)
\end{align}

We can then calculate contact force considering three cases for a single contact: opening,  stick, and  slip. In the case of opening contact, $v^n_{i,k+1} \geq 0$ even without contact force, the contact force is zero.
In the case of stick contact, the contact force can be obtained by utilizing the Signorini condition \eqref{eq:Signorini} and the condition that the tangential velocity at the contact node is zero, which is $v^t_{i,k+1} = \vec{t}_{i,k}^T J_{i,k} V_{k+1} = 0$. In the case of slip contact, we should solve the optimization of \eqref{eq:opt} for maximum energy dissipation together with Signorini condition \eqref{eq:Signorini} and Coulomb friction cone condition \eqref{eq:Coulomb}. 
This can be solved by using the bisection method \cite{HwangboRAL18} for single contact, and, for the multi-point contacts, we can extend that by using the nonlinear block successive over-relaxation (SOR) method \cite{OrtegaSIAM20}, which we adopt for our simulation framework here. 

\section{Configuration Space Contact Simulation}
\label{sec:Cspace}
\subsection{Contact Detection using Neural Network}
\label{subsec:CD_Cspace}
For contact detection in the C-space, we model the surface of the collision volume in SE(3) as a function of $q$, which can be written as $h(q) = 0 \in \Re$. This surface can be modeled by using the penetration depth similar to that for the N-space in Sec. 3.1. Here, unlike the N-space contact detection model, defining the C-space penetration depth is not straightforward, since it is non-trivial to come up with a proper distance metric combining both the translation and rotation of the object \cite{ZhangRSS07}. 
Instead, we construct the surface function $h(q)$ with a NN, which can be trained by using the collision-on and collision-off data by assigning a value of $h(q) = 1$ for the configuration without contact and $h(q) = -1$ for the configuration with contact.  The surface function $h(q)$ then has zero value on the contact surface, positive value for collision-free configuration, and negative value for collision configuration. For collecting data of configuration with/without collision, we can then calculate the penetration depth for all vertices of the object for various configurations and classify them as collision-free configurations if there are no contact nodes, and collision configuration otherwise. See \cite{SonIROS20} for more details.

\subsection{Contact Constraint in C-Space}
\label{subsec:CC_Cspace}
To prevent any part of the object from moving in the direction of increasing penetration, we formulate the contact constraint in the C-space similar to the velocity-level Signorini condition \eqref{eq:Signorini} in Sec. 2 s.t.,
\begin{align}
\label{eq:constraint_Cspace}
0 \leq \lambda^C_k \perp v^C_{k+1} \geq 0
\end{align}
where $\lambda^C_k \in \Re$  is the virtual contact force in the C-space (see Sec. 4.3), $v^C_{k+1} := (N^C_k)^T V_{k+1} \in \Re$ is the velocity along the C-space normal direction at next time index, and $N^C_k \in \Re^6$ is the C-space normal, which is the direction of the wrench that can move the object to the nearest point on the contact surface in the C-space from the current configuration. 
To determine this direction $N^C_k$, we should first find the velocity $V_C = \left[v_C;w_C\right] \in \Re^6$ moving from the current configuration to the nearest point on the contact surface, which can be obtained through the following optimization problem:
\begin{align*}
V_{C,\text{opt}} = \argmin_{V_C} \frac{1}{2} V_C^T M V_C
\end{align*}
subject to $h(q^C_k) = 0$, where $q^C_k$ is the configuration reached when moving at the velocity $V_C$ from the current configuration, which can be easily calculated from the kinematics. This optimization problem can be solved by using sequential quadratic programming (SQP). We can then find the force direction $N^C_k$ considering constraints related to dynamics, which is given by 
\begin{align*}
(N^C_k)^T M^{-1} \text{null}(V_{C,\text{opt}}) = 0
\end{align*}

\subsection{Integration with N-space Contact Model}
\label{subsec:Integ_Cspace}
\begin{figure}
	\centering
	\includegraphics[width=0.485\textwidth]{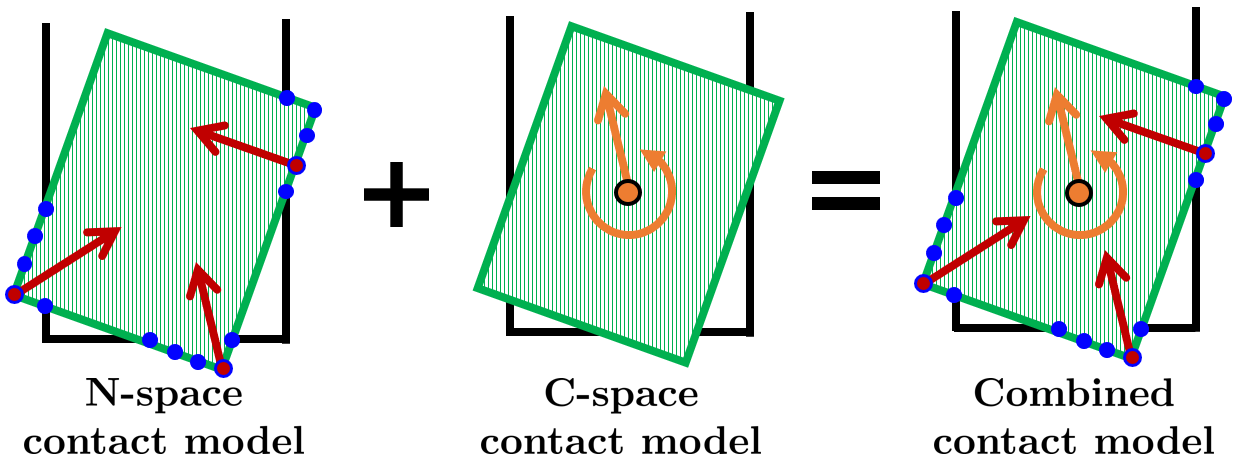}
	\caption{Illustration of N/C-combined contact solving: N-space contact model generating contact force for clustered contact nodes (left); C-space contact model generating virtual contact wrench for the object (middle); the combined contact model improve robustness of contact simulation (right).}
	\label{fig:Combined_CM}
    \vspace{-0.15in}
\end{figure}
To improve the robustness of the contact simulation while ensuring its accuracy, we combine the contact constraint in the C-space \eqref{eq:constraint_Cspace} and the contact model in the N-space in Sec. 3. For this, we add the virtual contact force $\lambda^C_k$ acting on the center of the object generated in the direction of reducing collision. This combined contact model is depicted in Fig. \ref{fig:Combined_CM}. Then, we modify the original dynamics equation  \eqref{eq:dynamics} s.t.,
\begin{align*}
M \frac{V_{k+1} - V_k}{T_k} + C_k(\omega_k) \frac{V_{k+1} + V_k}{2} = \sum_{i=1}^{M_c} J_{i,k}^T f_{i,k} + N^C_k \lambda^C_k
\end{align*}
where $N^C_k$ is the virtual wrench direction as defined in Sec. 4.2.

To solve the contact problem including the C-space contact model, we utilize the nonlinear block SOR method. The procedure is summarized in Algorithm \ref{algorithm:calculating_contact_force} and details are as follows. First, we compute the $i$-th contact force using the bisection method as described in Sec. 3.3 (line 4). Here, we use a successive relaxation form, $\lambda \leftarrow \alpha \lambda^* + (1-\alpha) \lambda$, where $\lambda$ is the current estimated solution, $\lambda^*$ is the optimal solution assuming that the other coupled variables are fixed, and $\alpha$ is a relaxation factor (line 5, 13). After all the calculation of the contact force in N-space is completed, we calculate the velocity $v^C_{k+1}$ without virtual contact force (line 7). Then, a virtual contact force is generated so that $v^C_{k+1}$ becomes zero if the object moves in the direction, in which more penetration increases (line 8, 9). Otherwise, we generate zero virtual contact force (line 10, 11). This process is repeated until the solution converges with the convergence determined by computing whether the contact constraints are satisfied. The total contact force is then given by
\begin{align*}
F_{c,k} =  \sum_{i=1}^{M_c} \Big(J_{i,k}^T (\vec{n}_{i,k} \lambda^n_{i,k} + \vec{t}_{i,k} \lambda^t_{i,k})\Big) + N^C_k \lambda^C_k \in \Re^6
\end{align*}
By combining the N-space clustered contact nodes/normals in Sec. 3.2 and the C-space contact constraint in this way, we can obtain the accurate contact force while improving the robustness against penetration for the entire part of the object. 

\begin{algorithm}[!t]
\SetAlgoLined
\KwResult{$\lambda_{1,k}^n,\lambda_{1,k}^t,\cdots,\lambda_{M_c,k}^n,\lambda_{M_c,k}^t,\lambda^C_k$}
	Initialize all variables to zero\\
	\While{not converged}{
		\For{$i \gets 1$  \KwTo $M_c$} {%
			$\Big((\lambda^n_{i,k})^*,(\lambda^t_{i,k})^*\Big)$ $\leftarrow$ SolveSingleContact\\
			$\begin{pmatrix}
			\lambda^n_{i,k}\\
			\lambda^t_{i,k}
			\end{pmatrix} 
			\leftarrow
			\alpha
			\begin{pmatrix}
			(\lambda^n_{i,k})^*\\
			(\lambda^t_{i,k})^*
			\end{pmatrix}
			+
			(1-\alpha)
			\begin{pmatrix}
			\lambda^n_{i,k}\\
			\lambda^t_{i,k}
			\end{pmatrix}$
		}
		compute $(v^C_{k+1})^*$ when $\lambda^C_k = 0$\\
		\eIf{$(v^C_{k+1})^* < 0$}{
			compute $(\lambda^C_k)^*$ using the equation $v^C_{k+1} = 0$
		}{
			$(\lambda^C_k)^* \leftarrow 0$
		}
		$\lambda^C_k \leftarrow \alpha (\lambda^C_k)^* + (1-\alpha) \lambda^C_k$
	}
	\caption{Calculating contact force at time step $k$}
	\label{algorithm:calculating_contact_force}
\end{algorithm}

\section{Experimental Validation}
\label{sec:Results}
\subsection{Simulation and Experimental Setups}
\label{sec:setup}
\begin{figure}
	\centering
	\includegraphics[width=0.485\textwidth]{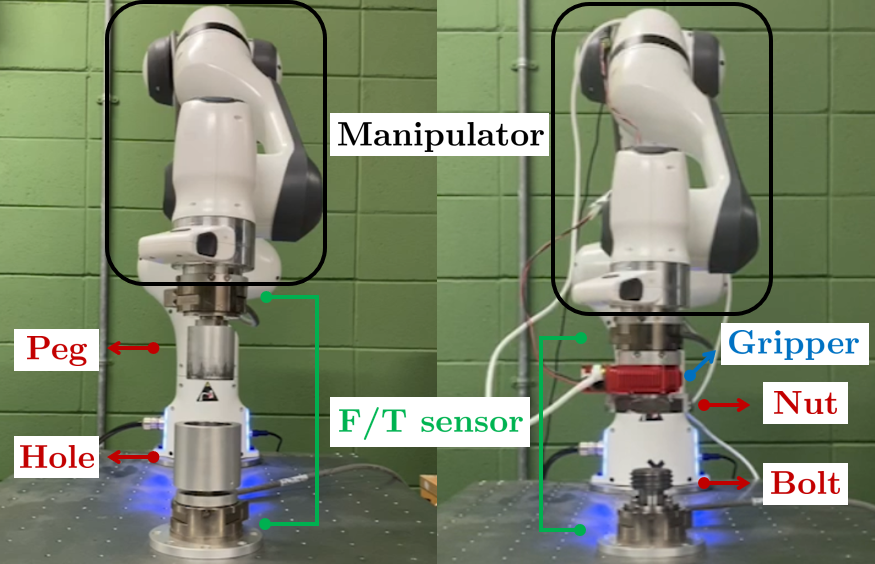}
	\caption{Experiment setup consisting of a 7-DOF robot arm and two F/T sensors (at the wrist and on the table): peg and hole for peg-in-hole task (left); and nut, bolt and infinite-turn gripper for bolting task (right). The two F/T sensors measure the contact wrench for learning and validation.}
	\label{fig:exp}
	\vspace{-0.15in}
\end{figure}

We conduct experiments to collect various data for training the contact clustering networks in Sec. 3.2. 
For this, as shown in Fig. \ref{fig:exp},  we configure the setup and perform the peg-in-hole task and the bolting task with tight machining tolerance (0.065mm for peg-in-hole, 0.4mm for bolting). 
The peg and hole are made of aluminum and have diameters of 49.97mm and 50.035mm, respectively.
The bolt and nut are made according to the M48 specification of the standard KS-B-0201 \cite{KSB} with the nominal $D=48$mm diameter and $p=5$mm pitch. 
Both the peg-in-hole and the bolting tasks are performed using a 7-DOF robot am (FRANKA EMIKA Panda) and two ATI Gamma F/T sensors, which are attached at the robot wrist and on the table to measure the contact wrenches between the robot arm and the peg/nut and that between the holed-cylinder/bolt and the grounding table. 
Other kinematic data (position, orientation, linear/angular velocity) are collected from the sensing values of the robot manipulator. For bolting task, we use an additional gripper (X-series of HEBI robotics) to rotate nut infinitely.

To collect rich data, a human user freely perform the two tasks by tele-controlling the robot arm with a haptic device connected via the standard virtual coupling \cite{KimRobotica17}.
The total data obtained are 190,000 simulation-steps for the peg-in-hole task and 330,000 simulation-steps for the bolting task.
We use 70$\%$ of these total experimental data to train the contact clustering networks and remaining 30$\%$ data as a test data for verification of the proposed simulation framework.

To validate our simulation framework, we then construct the simulation setup the same as that of the experiment in Fig. \ref{fig:exp}.
For this, we also include the generalized-coordinate PMI simulation of the robot manipulator with that of the object (i.e., peg or nut) in \eqref{eq:dynamics} according to \cite{KimIJRR17}:
\begin{gather*}
M_q(q_{r,k})\ddot{q}_{r,k} + C_q(q_{r,k},\dot{q}_{r,k})\frac{\dot{q}_{r,k+1}+\dot{q}_{r,k}}{2} = \tau_k + J_{q,k}^T \lambda_k\\
M \frac{V_{k+1} - V_k}{T_k} + C_k \frac{V_{k+1} + V_k}{2} = J_{X,k}^T \lambda_k + F_{c,k}
\end{gather*}
where $M_q \in \Re^{n_r \times n_r}, C_q \in \Re^{n_r \times n_r}$ are the inertia and Coriolis matrices of the robot arm, $q_{r,k} \in \Re^{n_r}$ is its joint angles with $n_r$ being its degree-of-freedom, $\tau_k \in \Re^{n_r}$ is the control torque, and $J_{q,k} \lambda_k \in \Re^{n_r}, J_{X,k}^T \lambda_k \in \Re^6$ are the potential action to enforce the kinematic constraint between the end-effector of the robot arm and the objects. See \cite{KimIJRR17} for more details. 
Here, for the bolting task, we add the infinite-turn gripper as the last joint of the robot manipulator (i.e., $n_r=7$ for the peg-in-hole task, $n_r = 8$ for the bolting task). 
To overcome non-negligible friction and backlash in the real robot manipulator, we adopt the admittance control \cite{Spong20} for $\tau_k$. We also ignore the gravity in simulation because we compensate the gravity using the robot manipulator in the experiment.

All the inertial parameters required for the simulation are experimentally identified using the technique of \cite{LeeTRO19} along with the friction coefficient. 
The poses of the static objects (i.e., holed-cylinder and bolt) required for the simulation are also experimentally identified as the penetration-minimizing poses when the peg is inserted more than a certain part into the hole or when the nut is tightened to the bolt more than a certain part, since the penetration would not occur (or be very small) in the real world.

\subsection{Design of Network Architectures}
\label{sec:architecture}

We construct the contact detection NNs for the N-space (Sec. 3.1) and for the C-space (Sec. 4.1) using MLP. 
The N-space contact detection MLP contains one input layer with a rectified linear unit (ReLU), two hidden layers with its size being sixty-four with ReLU, and one output layer with a hyperbolic tangent. The C-space contact detection MLP contains one input layer with a hyperbolic tangent, two hidden layers with its size being sixty-four with a hyperbolic tangent, and one output layer with a sigmoid function.

For the contact clustering INs/NNs (Sec. 3.2), we design the functions $f_{R_1},f_{R_2},f_{O_1},f_{O_2}$ using MLP.
The MLPs for $f_{R_1},f_{R_2}$ contain one input layer, two hidden layers with its size being sixteen, and one output layer with its size being four, all using ReLU as an activation function. 
The MLPs for $f_{O_1},f_{O_2}$ contain one input layer, one hidden layer with its size being eight, all using ReLU as an activation function, and one output layer that is a linear transform plus bias.
Here, to make the sum of weights to be one (i.e., $\sum_{i=1}^{N_c} W^j_i = 1, \forall j \in [1, M_c]$), the weights resulting from $f_{O_1}$ are normalized through the softmax activation function. 
For \eqref{eq:output}, the set $S_i$ is defined to be the set of ten closest points to the $i$-th contact node obtained by using the k-nearest neighbor search algorithm \cite{Nene97TPAMI}. 
We also set the number of the clustered points to be $M_c = 12$. All the inputs for NNs are also normalized by using the minimum and maximum values and rescaled to $[-1,1]$.

\subsection{Component-Wise Validation}
\label{sec:component}
\begin{table}
	\centering
	\caption{Computation time comparison of our proposed NN-based contact detection with a state-of-the-art/standard method of FCL \cite{PanICRA12}.}
	\label{tab:CD_perf}
	\vspace{0.05in}
	\begin{tabular}{|c||c|c|c|c|}
	\hline
	Task & \multicolumn{2}{c|}{Peg-in-Hole} & \multicolumn{2}{c|}{Bolting}\\
	\hline
	Method & FCL & Proposed & FCL & Proposed \\
	\hline
	\makecell{Average \\ time (ms)} & 2.31 & 0.28 & 5.82 & 0.28 \\
	\hline
	\makecell{Maximum \\ time (ms)} & 8.44 & 0.58 & 19.10 & 0.72 \\
	\hline
	\end{tabular}\\[10pt]
	\vspace{-0.15in}
\end{table}

Here, we validate some key components of our framework for the object-level simulation (i.e., peg or nut), while postponing that for the complete simulation framework with all the simulation pipelines of Fig. \ref{fig:diagram} implemented with the robot arm simulation also included in the next Sec. 5.4.

We first compare the computation time of our parallelized contact detection using NN in Sec. 3.1 with
FCL \cite{PanICRA12}, which is a standard contact detection method for general mesh objects.  
The evaluation is carried out by calculating the average and maximum computation time for the two tasks during total 94,000 simulation-steps for the peg-in-hole task and total 164,000 simulation-steps for the bolting task under various perturbations during the tasks.  
For the peg-in-hole task, the number of meshes for peg and the hole are 4,786 and 5,324, respectively, while, for the bolting task, that for bolt and the nut are 5,008 and 4,696, respectively. 
The results are shown in Table. \ref{tab:CD_perf}. There, we can see that our proposed method is about 8.2 times faster for the peg-in-hole task and 21.1 times faster for the bolting task than FCL based on the average computation time. This substantial speed-up is possible, since our proposed method requires only a simple calculation of the NN function values for only the contact nodes filtered by the broad phase contact detection process, which is further accelerated by parallelizing the computation to each contact node, unlike the exhaustive pair-wise test of typical standard methods.

\begin{figure}
\centering
	\includegraphics[width=0.465\textwidth]{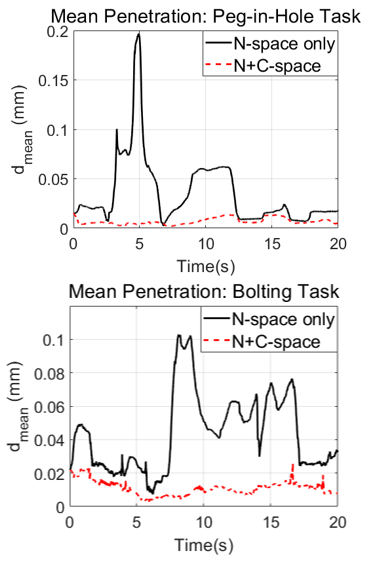}
	\caption{Mean penetration depth $|d_{\text{mean}}|$ of the N-space contact model only (black) and the combined N-space and C-space contact models (red) for the peg-in-hole task (top) and the bolting task (bottom).}
	\label{fig:Ceffect2}
	\vspace{-0.15in}
\end{figure}

We also evaluate efficacy of the inclusion of C-space contact model in Sec. 4. 
For this, we calculate the average of the penetration depth of all (original) contact nodes (i.e., object-level penetration) while applying a random disturbance wrench to the object (i.e., peg/nut) while being assembled with their counterparts (i.e., hole/bolt). 
This is also done with the complete simulation of Sec. 5.4 with all the pipelines of Fig. \ref{fig:diagram} and the robot arm simulation in Sec. 5.1.  
The results are shown in Fig. \ref{fig:Ceffect2}, where it is clear that, with the C-space contact model (Sec. 4)  combined with that of the (standard) N-space (Sec. 3), the penetration drastically reduces, since the C-space contact model in essence allows us to incorporate the penetration of all the (original) contact nodes $\mathcal{X}^o$, whereas the N-space contact model only those clustered contact nodes $\mathcal{X}^r$, thereby, significantly improve the robustness of contact simulation against penetration.

To verify the accuracy of contact force computation, and, consequently, the proper learning of the contact clustering networks (Sec. 3.2), we compare the simulated contact force (i.e., contact force computed
via the contact detection, contact clustering and contact solving, but with no time integration in Fig. \ref{fig:diagram} given the object state information from the test data from the experiment at one time index) with that of the experiment. 
For this, we consider only the objects and their simulation and not include the robot arm, since the aforementioned modules in Fig. \ref{fig:diagram} are developed for the objects, not for the robot arm.
For the complete evaluation with the robot arm, see instead Sec. 5.4. 
We also additionally implement the K-means clustering \cite{JainCSUR99} and the projected Gauss-Seidel (PGS \cite{Liu05TASE}) method in the places of our contact clustering and contact solving in Fig. \ref{fig:diagram} and compare their simulated contact force as well, since they can be thought of as a standard method for their respective role. 

\begin{figure}
	\centering
	\includegraphics[width=0.49\textwidth]{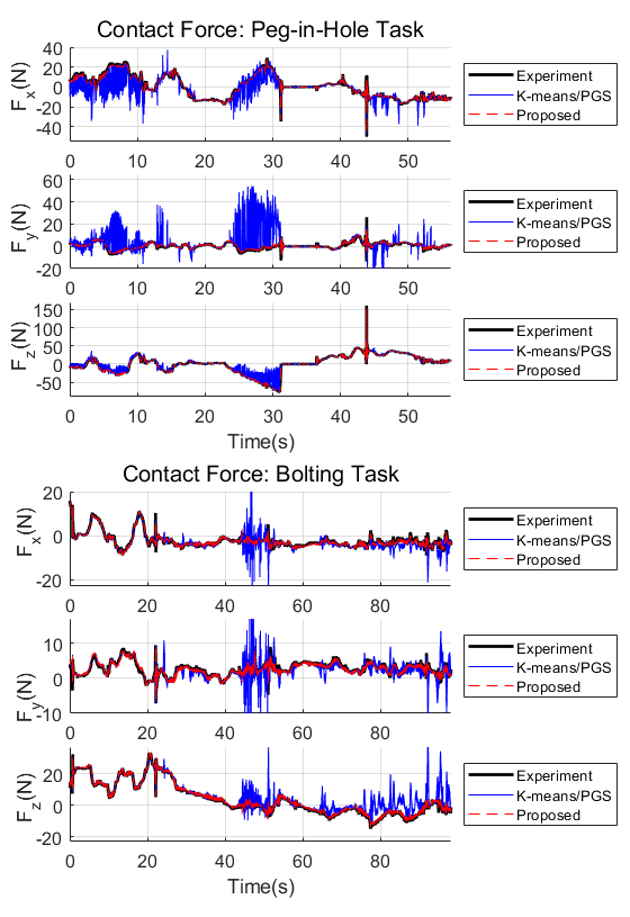}
	\caption{Simulated contact force of experiment and simulations with the standard K-mean/PGS method and our proposed framework for the peg-in-hole and bolting tasks.}
	\label{fig:comp}
	\vspace{-0.15in}
\end{figure}

The results are shown in Fig. \ref{fig:comp} for the peg-in-hole and bolting tasks,
where, for the standard K-means/PGS method, the contact nodes/normals suitable for the dynamic states of the object are not properly determined and the physical principles related to the contact are not accurately satisfied, causing a large contact force error w.r.t. the experiment data (root mean square error (RMSE) of the contact force: 16.27N for peg-in-hole task and 5.28N for bolting task). 
In contrast, our proposed framework can compute much accurate contact force as compared to the experimental contact force even though the number of contact nodes are reduced to a small number (RMSE of the contact force: 1.37N for peg-in-hole task and 0.72N for bolting task).

\subsection{Validation of Complete Simulation}
\label{subsec:RTSim}
\begin{figure*}
\centering
	\includegraphics[width=0.93\textwidth]{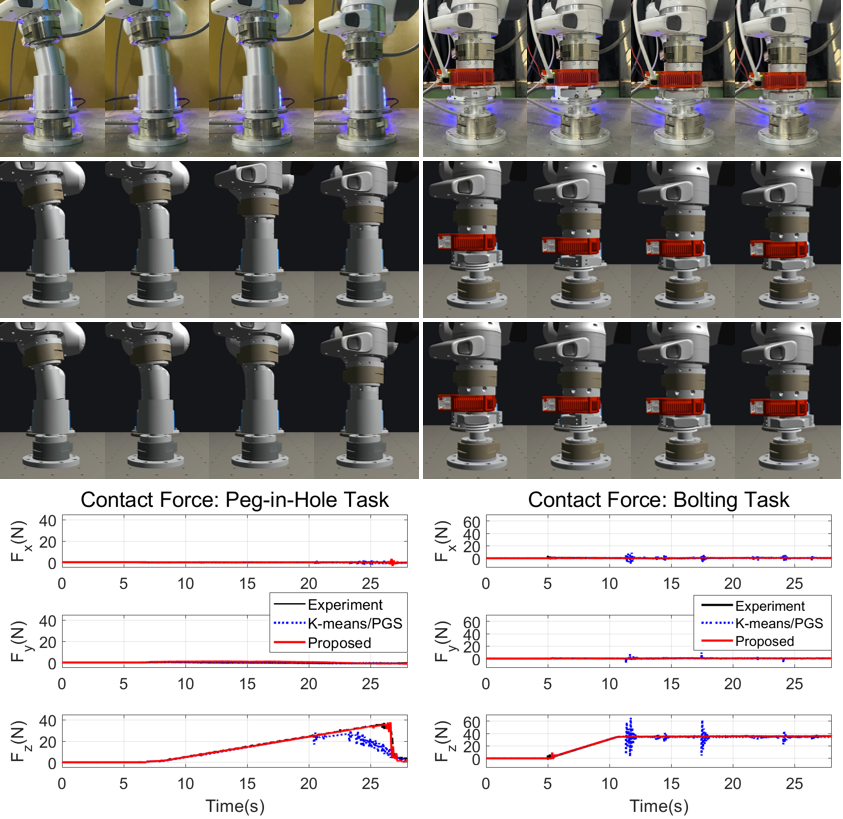}
	\caption{Snapshots of experiment (top row) and complete simulations with K-means/PGS (second row) and our proposed framework (third row) and their contact force plots of the peg-in-hole task (left) and the bolting task (right). }
	\label{fig:snapshot_and_force}
	\vspace{-0.05in}
\end{figure*}
\begin{figure}
	\centering
	\vspace{-0.2in}
	\includegraphics[width=0.43\textwidth]{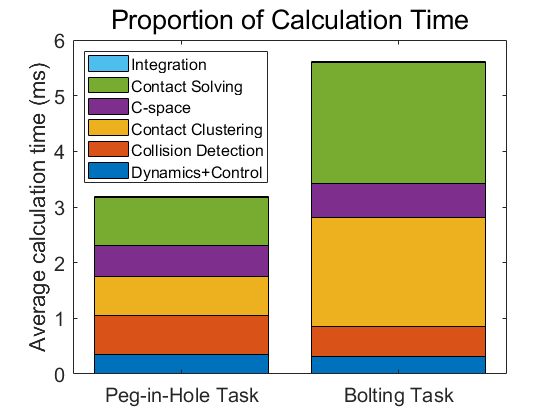}
	\caption{Averaged computation time for each pipeline during the complete simulation.}
	\label{fig:calc_time}
	\vspace{-0.15in}
\end{figure}
\begin{figure*}[ht]
\centering
	\includegraphics[width=0.95\textwidth]{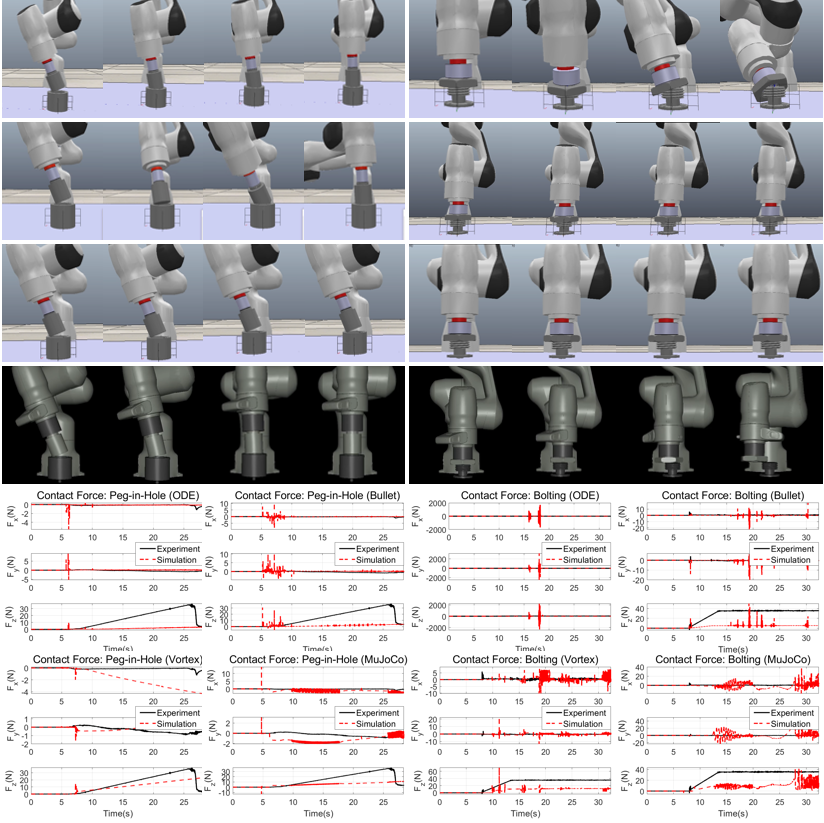}
	\caption{Snapshots of experiment and simulations with their contact force plots for the peg-in-hole task (left) and the bolting task (right) with open-source robot simulators: ODE (first row), Bullet (second row), Vortex (third row) and MuJoCo (fourth row).}
	\label{fig:snapshot_opensource}
	\vspace{-0.15in}
\end{figure*}

Here, we validate the performance and behavior of the complete simulation of our proposed framework, i.e., using all the pipelines of Fig. \ref{fig:diagram} with the robot arm simulation also included with the matching with the experimental setup data done only for the simulation initial condition. 
Snapshots and contact force plots of the experiment and the complete simulations with the standard K-means/PGS method (see Sec. 5.3) and our proposed framework for the peg-in-hole and bolting tasks are presented in Fig. \ref{fig:snapshot_and_force} and also in the accompanied movie, where we can see that our proposed framework fits very well with the experimental behavior and the contact force is very similar to the measured contact force in the experiment (RMSE of contact force: 1.63N for peg-in-hole task and 1.01N for bolting task), whereas the K-means/PGS method exhibits some discrepancy, particularly, sudden penetrating motions during the two tasks, which is fairly prominent from the movie file for the bolting task.

The average one-step computation times for the simulations are shown in Fig. \ref{fig:calc_time} with the breakdown for each simulation pipeline of Fig. \ref{fig:diagram} as well. 
In fact, the average computation time is around 3.2ms for the peg-in-hole task and 5.6ms for the bolting task, implying that our framework can real-time simulate these tasks, if we use 10ms as the integration-step, which turns out to be adequate for the scenarios reported here.
From these results, we can verify that both physically-accurate and real-time simulation is possible even for these challenging contact-intensive tight-tolerance assembly tasks by using our proposed simulation framework. 
This is, however, very hard (or often impossible) to attain by using open-source robot simulators.

To show this, we implement the peg-in-hole and bolting tasks scenarios with some representative open-source robot simulators (e.g., CoppeliaSim, MuJoCo) while changing its physics engine (e.g., ODE \cite{SmithODE05}, Bullet \cite{CoumansBULLET13}, Vortex \cite{Vortex}, MuJoCo \cite{TodorovMUJOCO12}). 
The results are shown in Fig. \ref{fig:snapshot_opensource}, where we found that it is very hard (if not impossible) to properly simulate such contact-intensive tight-tolerance tasks as the peg-in-hole and bolting tasks using the currently available open-source robot simulators, all of which exhibit substantial discrepancy from the real experiment data and, in some cases, even become unstable.
For the simulations in  Fig. \ref{fig:snapshot_opensource}, we match their simulation kinematic/inertial parameters to the experimental data as explained in Sec. 5.1, while, for other simulation parameters, particularly those for the contact solving (e.g., for MuJoCo), we adopt their suggested/default values or optimize them to match the experimental data on our own, although the overall behavior of the simulations do not change much from that in Fig. \ref{fig:snapshot_opensource} even with some modification of those identified parameters.

\subsection{Generalization}
\label{subsec:GP}
\begin{figure}
\centering
	\includegraphics[width=0.485\textwidth]{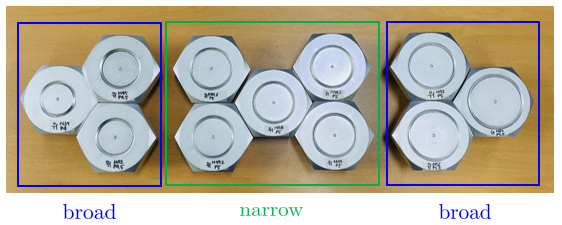}
	\caption{Bolts and nuts of various diameters/pitches for the narrow and broad generalization groups.}
	\label{fig:generalization_exp_setup}
\end{figure}

\begin{table}
	\centering
	\caption{Diameters and pitches of various bolts and nuts for generalization verification.}
	\label{tab:DP}
	\vspace{0.05in}
	\begin{tabular}{|c||c|c|c|c|c|c|}
	\hline
	 & \multicolumn{6}{c|}{Narrow generalization}\\
	\hline
	Diameter (mm) & 47.2 & 47.6 & 48.5 & 49 & - & - \\
	\hline
	Pitch (mm) & 5 & 5 & 5 & 5 & - & -\\	
	\hline
	 & \multicolumn{6}{c|}{Broad generalization}\\	
	\hline
	Diameter (mm)  & 39  & 42  & 45  & 52  & 56  & 60 \\
	\hline
	Pitch (mm) & 4 & 4.5 & 4.5 & 5 & 5.5 & 5.5 \\
	\hline
	\end{tabular}\\[10pt]
	\vspace{-0.15in}
\end{table} 

To evaluate the generalization capacity of our proposed simulation framework, we perform the bolting tasks using bolts/nuts with various perturbed geometries. 
More precisely, from the nominal $D=48$mm diameter and $p=5$mm pitch of KS-standard M48 bolt/nut \cite{KSB}, 
we define the narrow and broad generalization groups,
with the former perturbed for their diameters slightly larger than (or similar to) the manufacturing tolerance  (i.e., 0.335mm for bolt, 0.45mm for nut), whereas the latter much larger while similar to the tolerance for their pitches.
Four bolt-nut pairs are machined for the narrow group, whereas six pairs for the broad group - see Fig. \ref{fig:generalization_exp_setup} and Table \ref{tab:DP} for their diameters and parameters.
Here, we introduce the narrow generalization group to see if our simulation framework can cope with the machining tolerance of the bolt/nut, whereas the broad group to see if it can be generalized to even a larger extent (e.g., different specification bolts/nuts). 
Defining the values for Table \ref{tab:DP}, we also take into account the fact that manufacturing tolerance typically dominantly occur in the radial direction rather than the pitch direction for bolts and nuts.

To evaluate the generalization of our framework, we perform simulation for each of these ten pairs of (perturbed) bolts/nuts and compare their results with the experiment data. 
For this, we newly the contact detection NNs in Sec. 3.1 and 4.1 for each of them, which can be done with no experiments, since it is purely kinematic procedure, thus, can be off-line performed only using the CAD model data for each bolt/nut pair.  
In contrast, the contact clustering networks in Sec. 3.2 are not retrained at all and the networks trained in Sec. 5.2 for the nominal M48 bolt/nut with $D=48$mm and $p=5$mm are just applied to all the ten perturbed pairs in  Table \ref{tab:DP}. 

\begin{figure}
\centering
	\includegraphics[width=0.485\textwidth]{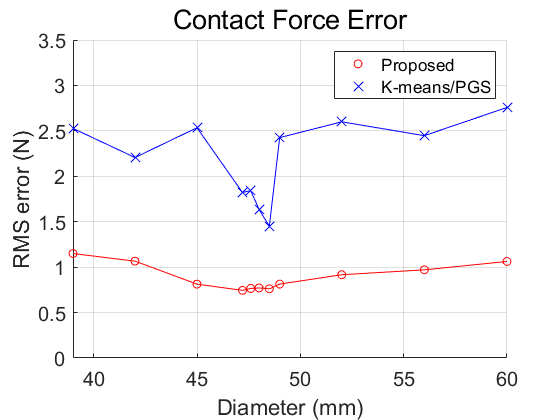}
	\caption{Generalization performance of our proposed simulate framework and the standard K-means/PGS method
		for the bolting task: RMSE of the norm of the contact force.}
	\label{fig:Contact_Force_Error}
	\vspace{-0.15in}
\end{figure}

\begin{figure*}
\centering
	\includegraphics[width=0.95\textwidth]{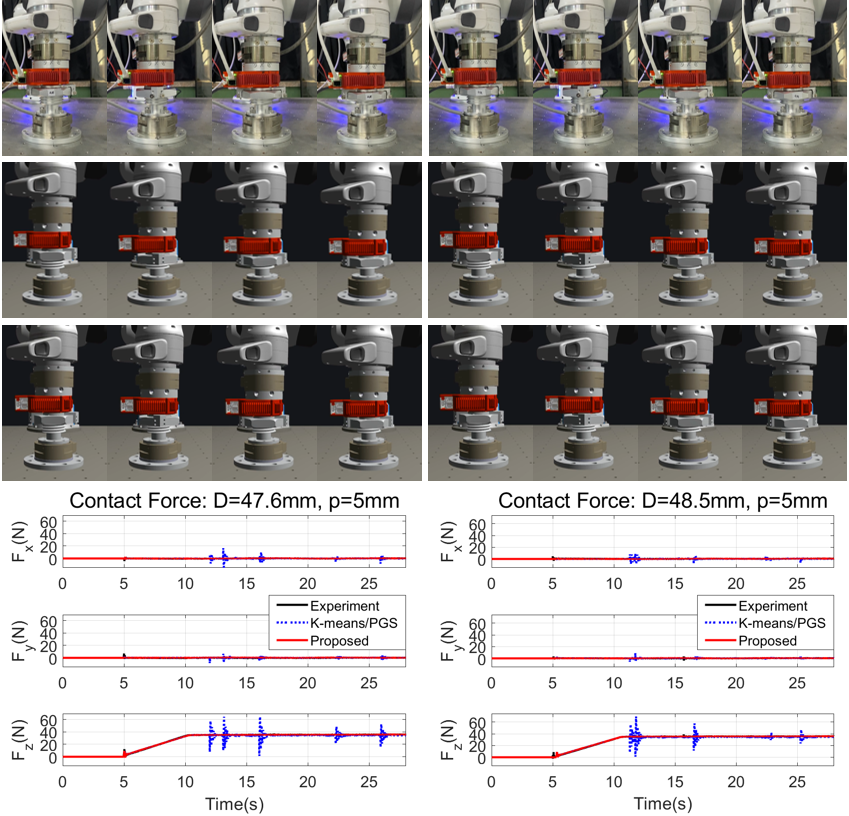}
	\caption{Snapshots of experiment and simulations with the K-means/PGS method and our proposed framework with their contact force plots for the narrow generalization performance verification.}
	\label{fig:snap_narrow}
\end{figure*}

\begin{figure*}
\centering
	\includegraphics[width=0.95\textwidth]{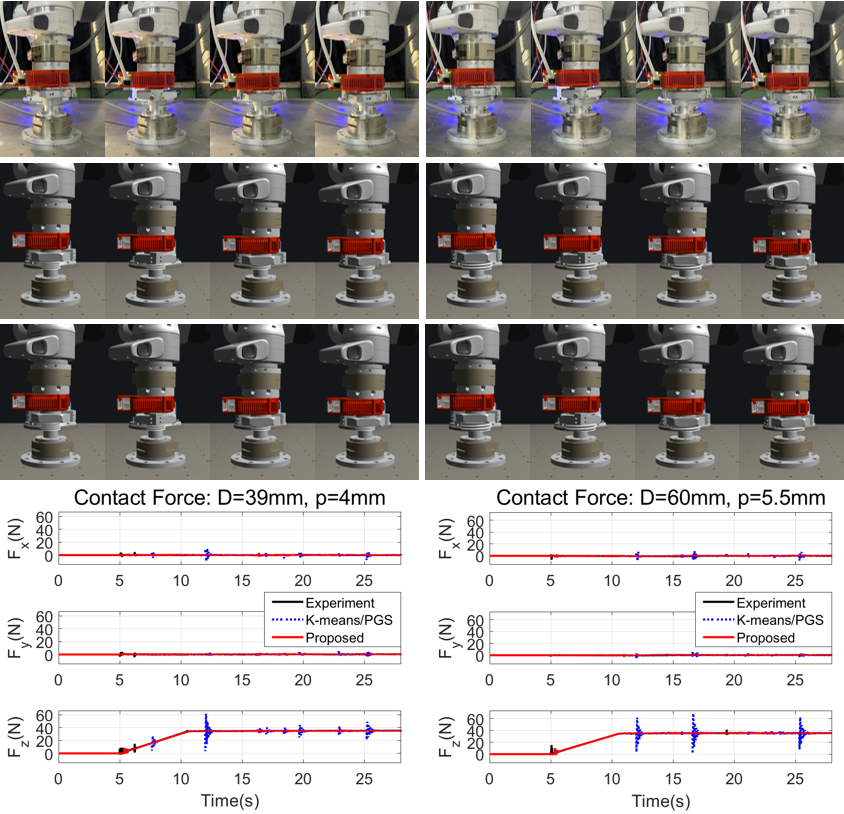}
	\caption{Snapshots of experiment and simulations with the K-means/PGS method and our proposed framework with their contact force plots for the broad generalization performance verification.}
	\label{fig:snap_broad}
	\vspace{-0.15in}
\end{figure*}

Error of the object simulated contact force (i.e., object contact force computed from the contact detection/clustering/solving at each simulation-step given the object states of the experiment with the robot arm - see Sec. 5.3 and Fig. \ref{fig:comp}) 
are shown in Fig. \ref{fig:Contact_Force_Error}, where we can clearly see that our proposed framework is indeed capable for generalization at least across the ten perturbed pairs. 
We also perform the complete simulation (i.e., using all the pipelines of Fig. \ref{fig:diagram} with the robot arm also included - see Sec. 5.4 and Fig. \ref{fig:snapshot_and_force}) of the ten perturbed pairs.
Snapshots of the experiments and simulations for two cases from each of the narrow and broad generalization groups are shown in Fig. \ref{fig:snap_narrow} and Fig. \ref{fig:snap_broad} (and also in the accompanied movie), where we can also clearly see that our proposed simulation framework can maintain its speed (all simulation speed is similar as reported in Fig. \ref{fig:calc_time}) and accuracy across the perturbed ten pairs, thereby, manifesting its generalization capacity at least to the extent similar to those as captured by those perturbed pairs. 
It is also interesting to see from Fig. \ref{fig:snap_narrow} and Fig. \ref{fig:snap_broad} (and the accompanied movie) that the perturbed pairs still exhibit simulation behaviors similar to that of the nominal M48 bolt/nut pair, including those sudden penetrating motions with K-means/PGS method, even if its error appears rather small from Fig. \ref{fig:Contact_Force_Error}. 

\subsection{Haptic Rendering of Virtual Bolting Task}
\label{subsec:App}

To show the robustness and versatility of our proposed simulation framework, we apply it to the problem of haptic rendering of virtual bolting task with arbitrary human commands. 
Our proposed simulation framework is particularly suitable for this, since: 1) it is accurate and fast, thus, should be able to believably reproduce the real-world behavior of the bolting task with interactively-fast update rate; and 2) from its relying on the discrete-time passive PMI \cite{KimIJRR17}, it can substantially enhance the interaction stability of the haptic rendering with human operators, while also being compatible to standard technqiues for haptic rendering/control such as four-channel transparent virtual coupling \cite{lee2012passive,kim2013toward} even with some delay  \cite{LeeTRO13}.

\begin{figure}
\centering
	\includegraphics[width=0.485\textwidth]{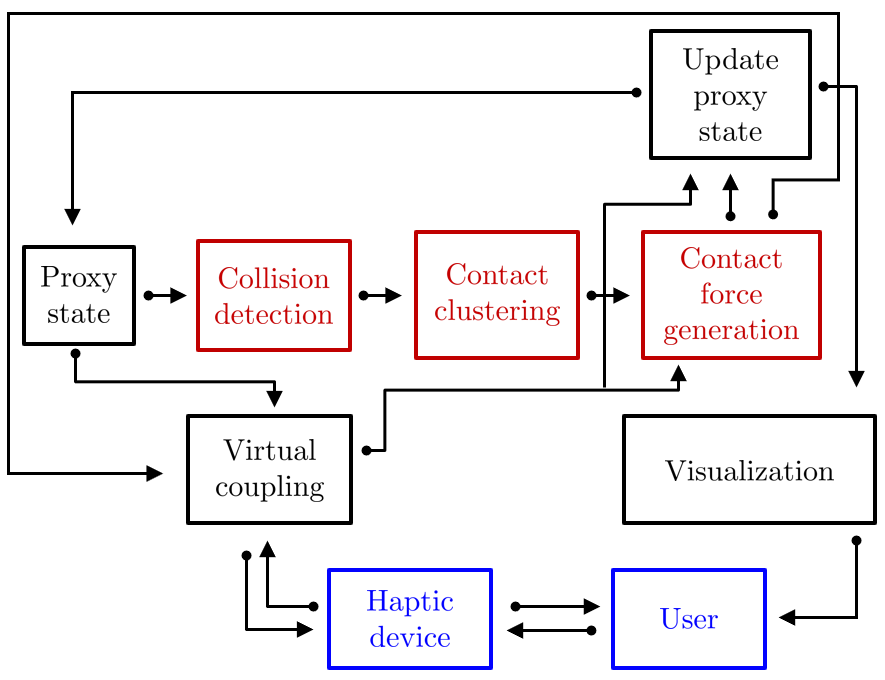}
	\caption{Schematic of haptic rendering of virtual bolting task, consisting of our proposed simulation pipelines for the bolting task, with the nut dynamics coupled with a haptic device via the transparent virtual coupling \cite{KimRobotica17,kim2013toward} and a visualization display.}
	\label{fig:tele}
\vspace{-0.15in}
\end{figure}

\begin{figure*}
\centering
	\includegraphics[width=\textwidth]{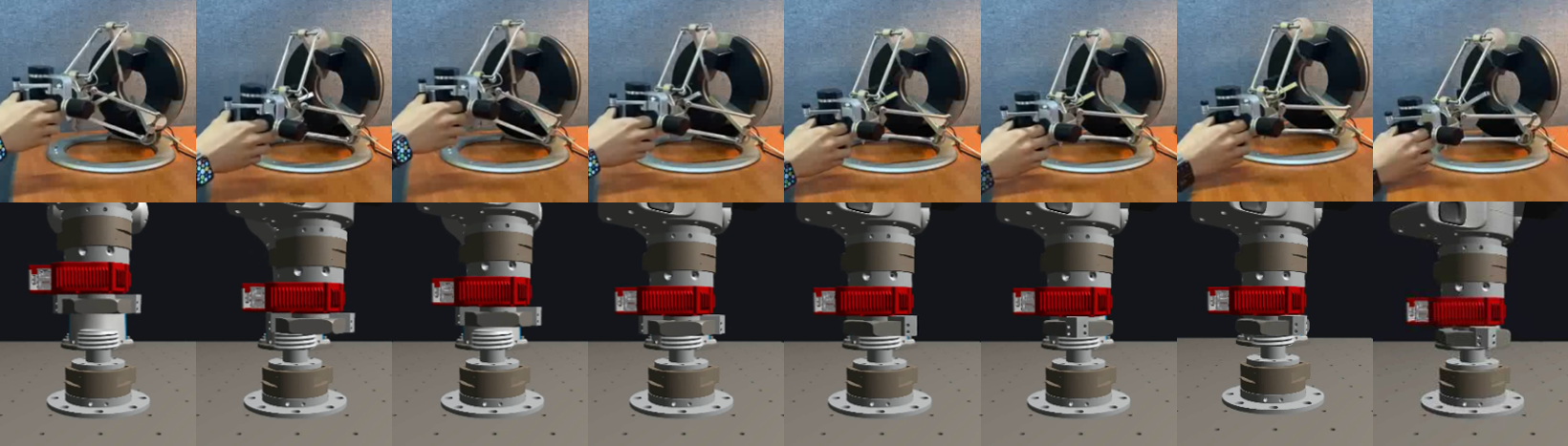}
	\caption{Snapshots of the haptic rendering of virtual bolting task with the human controlling the haptic device (top) and the behavior of the simulated nut, bolt and robot manipulator in the virtual world (bottom).}
	\label{fig:tele_fig4}
	\vspace{-0.15in}
\end{figure*}

\begin{figure}
\centering
	\includegraphics[width=0.485\textwidth]{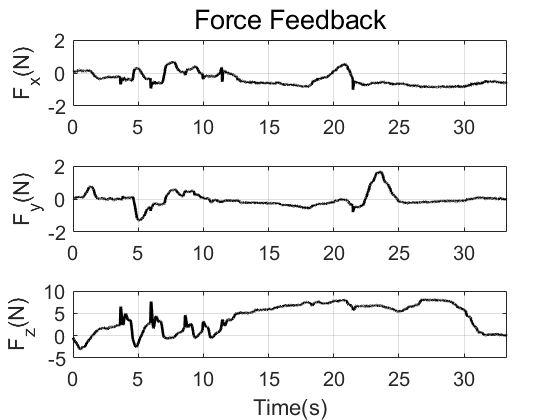}
	\caption{Force feedback to the user through the haptic device when performing the virtual bolting task. }
	\label{fig:tele_fig5}
	\vspace{-0.15in}
\end{figure}

Shown in Fig. \ref{fig:tele} is the overall architecture for the haptic rendering of virtual bolting task, which 
is composed of: 1) our proposed (real-time/accurate/passive) simulation pipelines of the virtual nut object (i.e., proxy); 2) four-channel transparent passivity-based virtual coupling \cite{KimRobotica17}, which connects the proxy in the virtual world to the haptic device configuration in the real world; and 3) haptic device, which transmits human command to the proxy while providing haptic feedback through the virtual coupling; and 4) visualization module, which displays simulated objects (e.g., bolt, nut, and manipulator) and their environment to the user. Here, the robot arm is also incorporate in the simulation and it is controlled with an admittance control with the proxy state in Fig. \ref{fig:tele} as its set pose of its end-effector.  
We also utilize keyboard input to rotate the nut.

Snapshots of this haptic rendering of the virtual bolting task are shown in Fig. \ref{fig:tele_fig4} with its contact force also presented in Fig. \ref{fig:tele_fig5}, where we can observe that the simulation behavior is physically reasonable and also robust even with arbitrary human command.
For instance, before around 12sec in Fig. \ref{fig:tele_fig5} (i.e., first four snapshots of Fig. \ref{fig:tele_fig4}), when the user initially tries to find the fastening position of the nut by moving around it around the upper surface of the bolt, the contact force (sometimes like an impulse) is generated to stop the nut and prevent penetration. 
After the fastening position is found, as shown from the right four snapshots in Fig. \ref{fig:tele_fig4} (or after around 12sec in Fig. \ref{fig:tele_fig5}), 
the nut is well maintained fastened to the bolt while giving reasonable force feedback to the user through the haptic device. 
Even if the human applies the command in any direction after the fastening, the nut is well maintained fastened without losing the contact or exhibiting penetration as in reality. 

\section{Conclusion and Future Work} \label{sec:conclusion}

In this paper, we propose a novel fast (i.e., faster than, or at least near, real-time) and accurate (i.e., matching well with experiment contact wrench data) simulation framework for the contact-intensive tight-tolerance tasks. For this, we develop the parallelized data-driven contact detection technique with neural network (NN) and the accuracy-optimized data-driven contact clustering technique with interaction network (IN), together, they can significantly improve the accuracy and speed of the simulation framework while able to efficiently accommodate objects with complicated and non-convex shapes.  
We also develop the constraint/energy-based contact solving technique, that can strictly enforce the key physical principles of contact (e.g., non-penetration constraint, maximum energy dissipation, Coulomb friction cone) even in the discrete-time domain, while also combining the contact model for the contact nodes and that of the object configuration, thereby, significantly enhancing the accuracy, stability and robustness of the simulation. 
We further formulate our simulation framework on the foundation of PMI \cite{KimIJRR17}, which, by enhancing passivity in the discrete-time domain, can further improve overall stability and speed of our simulation framework. 
Various experimental validations are also performed to verify the speed, accuracy, generalizability, robustness and versatility of our proposed simulation framework.

Some possible future research topics include: 
1) applications of the proposed simulation frameworks for other contact-intensive and tight-tolerance robotic rigid-object assembly scenarios;
2) applications for data-driven/RL-based sim-to-real control policy developments for the contact-intensive tight-tolerance tasks, including the peg-in-hole and bolting tasks as considered in this paper; and
3) theoretical extension to the manipulation applications of non-rigid and deformable objects.


\begin{IEEEbiography}
[{\includegraphics[width=1.05in,height=5in,clip,keepaspectratio]{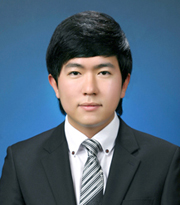}}]{\bf Jaemin Yoon} received the B.S. degree and Ph.D. degree in Mechanical $\&$ Aerospace Engineering from the Seoul National University, Seoul, Republic of Korea, in 2013 and 2021, respectively. He is now with the Robot Center, Samsung Research, Seoul, Republic of Korea. His main research interests include rigid/soft object and robot simulation, machine learning, mobile manipulator, and object manipulation.
\end{IEEEbiography}

\begin{IEEEbiography}
[{\includegraphics[width=1.05in,height=5in,clip,keepaspectratio]{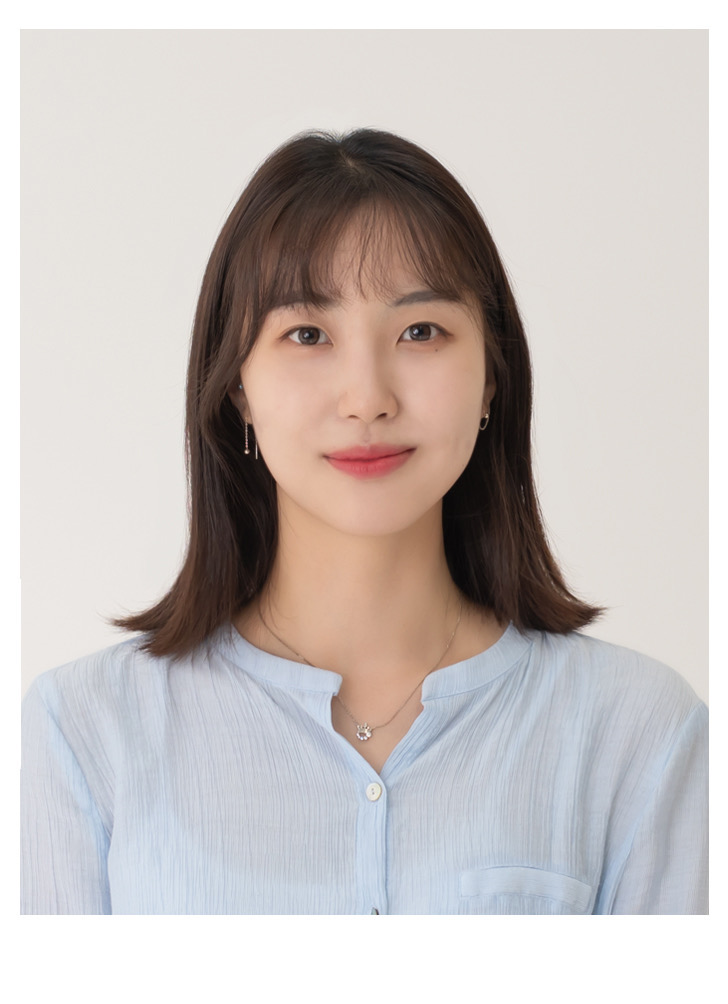}}]{\bf Minji Lee} received the B.S. degree in mechanical engineering in 2019 from Seoul National University, Seoul, Republic of Korea, where she is currently working toward the Ph.D. degree in mechanical engineering. Her main research interests include the control-theoretic simulation of soft robots with multi-contact.
\end{IEEEbiography}

\begin{IEEEbiography}
[{\includegraphics[width=1.05in,height=5in,clip,keepaspectratio]{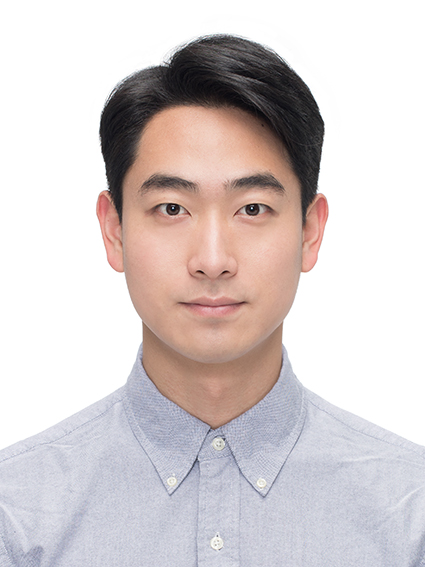}}]{\bf Dongwon Son} received the B.S. degree and Master degree in Mechanical $\&$ Aerospace Engineering from the Seoul National University, Seoul, Republic of Korea, in 2014 and 2020, respectively. He is now with the AI Method Team, Samsung Research, Seoul, Republic of Korea. His main research interests include robot simulation, object manipulation, and machine learning.
\end{IEEEbiography}

\begin{IEEEbiography}
[{\includegraphics[width=1.05in,height=5in,clip,keepaspectratio]{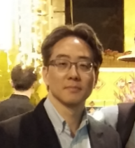}}]{\bf Dongjun Lee} received the B.S. degree in Mechanical Engineering from KAIST, Daejeon, the M.S. degree in Automation $\&$ Design from KAIST, Daejeon, and the Ph.D. degree in Mechanical Engineering from the University of Minnesota at Twin Cities, 2004. He is currently a Professor with the Department of Mechanical Engineering, Seoul National University, Seoul, Republic of Korea. His main research interests include dynamics and control of robotic and mechatronic systems with emphasis on teleoperation/haptics, aerial robots, multirobot systems and industrial control applications.
\end{IEEEbiography}


\begin{thebibliography}{10}
\providecommand{\url}[1]{#1}
\csname url@samestyle\endcsname
\providecommand{\newblock}{\relax}
\providecommand{\bibinfo}[2]{#2}
\providecommand{\BIBentrySTDinterwordspacing}{\spaceskip=0pt\relax}
\providecommand{\BIBentryALTinterwordstretchfactor}{4}
\providecommand{\BIBentryALTinterwordspacing}{\spaceskip=\fontdimen2\font plus
\BIBentryALTinterwordstretchfactor\fontdimen3\font minus
  \fontdimen4\font\relax}
\providecommand{\BIBforeignlanguage}[2]{{%
\expandafter\ifx\csname l@#1\endcsname\relax
\typeout{** WARNING: IEEEtran.bst: No hyphenation pattern has been}%
\typeout{** loaded for the language `#1'. Using the pattern for}%
\typeout{** the default language instead.}%
\else
\language=\csname l@#1\endcsname
\fi
#2}}
\providecommand{\BIBdecl}{\relax}
\BIBdecl

\bibitem{KimIJRR17}
M.~Kim, Y.~Lee, Y.~Lee, and D.~J. Lee, ``Haptic rendering and interactive
  simulation using passive midpoint integration,'' \emph{Int. J. Robot. Res.},
  vol.~36, no.~12, pp. 1341--1362, 2017.

\bibitem{WangIJAC18}
T.~M. Wang, Y.~Tao, and H.~Liu, ``Current researches and future development
  trend of intelligent robot: A review,'' \emph{Int. J. Autom. Comput.},
  vol.~15, no.~5, pp. 525--546, 2018.

\bibitem{SanchezIJRR18}
J.~Sanchez, J.~A. Corrales, B.~C. Bouzgarrou, and Y.~Mezouar, ``Robotic
  manipulation and sensing of deformable objects in domestic and industrial
  applications: a survey,'' \emph{Int. J. Robot. Res.}, vol.~37, no.~7, pp.
  688--716, 2018.

\bibitem{ParkTIE17}
H.~Park, J.~Park, D.~H. Lee, J.~H. Park, M.~H. Baeg, and J.~H. Bae,
  ``Compliance-based robotic peg-in-hole assembly strategy without force
  feedback,'' \emph{IEEE Trans. Ind. Electron.}, vol.~64, no.~8, pp.
  6299--6309, 2017.

\bibitem{XuAccess19}
J.~Xu, C.~Zhang, Z.~Liu, and Y.~Pei, ``A review on significant technologies
  related to the robot-guided intelligent bolt assembly under complex or
  uncertain working conditions,'' \emph{IEEE Access}, vol.~7, pp.
  136\,752--136\,776, 2019.

\bibitem{MJ_IROS21}
M.~Lee, J.~Lee, J.~Yoon, and D.~J. Lee, ``Real-time physically-accurate
  simulation of robotic snap connection process,'' in \emph{IEEE/RSJ Int. Conf.
  Intell. Robots Syst.}, 2021, pp. 5173--5180.

\bibitem{SonIROS20}
D.~Son, H.~Yang, and D.~J. Lee, ``Sim-to-real transfer of bolting tasks with
  tight tolerance,'' in \emph{IEEE/RSJ Int. Conf. Intell. Robots Syst.}, 2020,
  pp. 9056--9063.

\bibitem{levine2020offline}
S.~Levine, A.~Kumar, G.~Tucker, and J.~Fu, ``Offline reinforcement learning:
  Tutorial, review, and perspectives on open problems,''
  \emph{arXiv:2005.01643}, 2020.

\bibitem{LevineIJRR18}
S.~Levine, P.~Pastor, A.~Krizhevsky, J.~Ibarz, and D.~Quillen, ``Learning
  hand-eye coordination for robotic grasping with deep learning and large-scale
  data collection,'' \emph{Int. J. Robot. Res.}, vol.~37, no. 4-5, pp.
  421--436, 2018.

\bibitem{Vortex}
CMLABS, ``Vortex studio,'' \emph{http://www.cm-labs.com}, 2017.

\bibitem{SmithODE05}
R.~Smith \emph{et~al.}, ``Open dynamics engine,'' 2005.

\bibitem{CoumansBULLET13}
E.~Coumans \emph{et~al.}, ``Bullet physics library,'' \emph{Open source:
  bulletphysics.org}, vol.~15, no.~49, p.~5, 2013.

\bibitem{TodorovMUJOCO12}
E.~Todorov, T.~Erez, and Y.~Tassa, ``Mujoco: A physics engine for model-based
  control,'' in \emph{IEEE/RSJ Int. Conf. Intell. Robots Syst.}, 2012, pp.
  5026--5033.

\bibitem{JainCSUR99}
A.~K. Jain, M.~N. Murty, and P.~J. Flynn, ``Data clustering: a review,''
  \emph{ACM Comput. Surveys}, vol.~31, no.~3, pp. 264--323, 1999.

\bibitem{BattagliaANIPS16}
P.~W. Battaglia, R.~Pascanu, M.~Lai, D.~Rezende, and K.~Kavukcuoglu,
  ``Interaction networks for learning about objects, relations and physics,''
  \emph{arXiv:1612.00222}, 2016.

\bibitem{KimICRA19}
M.~Kim, J.~Yoon, D.~Son, and D.~J. Lee, ``Data-driven contact clustering for
  robot simulation,'' in \emph{IEEE Int. Conf. Robot. Autom.}, 2019, pp.
  8278--8284.

\bibitem{DrumTVCG08}
E.~Drumwright, ``A fast and stable penalty method for rigid body simulation,''
  \emph{IEEE Trans. Vis. Comput. Graphics}, vol.~14, no.~1, pp. 240--231, 2008.

\bibitem{otaduy2006modular}
M.~A. Otaduy and M.~C. Lin, ``A modular haptic rendering algorithm for stable
  and transparent 6-dof manipulation,'' \emph{IEEE Trans. Robot.}, vol.~22,
  no.~4, pp. 751--762, 2006.

\bibitem{lloyd2005fast}
J.~E. Lloyd, ``Fast implementation of lemke's algorithm for rigid body contact
  simulation,'' in \emph{IEEE Int. Conf. Robot. Autom.}, 2005, pp. 4538--4543.

\bibitem{HorakRAL19}
P.~C. Horak and J.~C. Trinkle, ``On the similarities and differences among
  contact models in robot simulation,'' \emph{IEEE Robot. Autom. Lett.},
  vol.~4, no.~2, pp. 493--499, 2019.

\bibitem{StewartSIAM20}
D.~E. Stewart, ``Rigid-body dynamics with friction and impact,'' \emph{SIAM
  Review}, vol.~42, no.~1, pp. 3--39, 2000.

\bibitem{PreclikCM18}
T.~Preclik, S.~Eibl, and U.~R{\"u}de, ``The maximum dissipation principle in
  rigid-body dynamics with inelastic impacts,'' \emph{Comput. Mech.}, vol.~62,
  no.~1, pp. 81--96, 2018.

\bibitem{GilbertJRA88}
E.~G. Gilbert, D.~W. Johnson, and S.~S. Keerthi, ``A fast procedure for
  computing the distance between complex objects in three-dimensional space,''
  \emph{IEEE J. Robot. Autom.}, vol.~4, no.~2, pp. 193--203, 1988.

\bibitem{SnethenGPG08}
G.~Snethen, ``Xenocollide: Complex collision made simple,'' in \emph{Game
  programming Gems 7}.\hskip 1em plus 0.5em minus 0.4em\relax Course
  Technology, 2008, pp. 165--178.

\bibitem{PanICRA12}
J.~Pan, S.~Chitta, and D.~Manocha, ``Fcl: A general purpose library for
  collision and proximity queries,'' in \emph{IEEE Int. Conf. Robot. Autom.},
  2012, pp. 3859--3866.

\bibitem{atkinsonJWS08}
K.~E. Atkinson, \emph{An introduction to numerical analysis}.\hskip 1em plus
  0.5em minus 0.4em\relax John wiley \& sons, 2008.

\bibitem{butcherWOL08}
J.~C. Butcher and N.~Goodwin, \emph{Numerical methods for ordinary differential
  equations}.\hskip 1em plus 0.5em minus 0.4em\relax Wiley Online Library,
  2008, vol.~2.

\bibitem{hairerAN03}
E.~Hairer, C.~Lubich, and G.~Wanner, ``Geometric numerical integration
  illustrated by the st{\"o}rmer-verlet method,'' \emph{Acta Numerica},
  vol.~12, pp. 399--450, 2003.

\bibitem{haykin2010neural}
S.~Haykin, \emph{Neural networks and learning machines, 3/E}.\hskip 1em plus
  0.5em minus 0.4em\relax Pearson Education India, 2010.

\bibitem{rustamovCGF09}
R.~M. Rustamov, Y.~Lipman, and T.~Funkhouser, ``Interior distance using
  barycentric coordinates,'' \emph{Comput. Graphics Forum}, vol.~28, no.~5, pp.
  1279--1288, 2009.

\bibitem{sak2014long}
H.~Sak, A.~W. Senior, and F.~Beaufays, ``Long short-term memory recurrent
  neural network architectures for large scale acoustic modeling,'' 2014.

\bibitem{KernNC04}
S.~Kern, S.~D. M{\"u}ller, N.~Hansen, D.~B{\"u}che, J.~Ocenasek, and
  P.~Koumoutsakos, ``Learning probability distributions in continuous
  evolutionary algorithms--a comparative review,'' \emph{Nat. Comput.}, vol.~3,
  no.~1, pp. 77--112, 2004.

\bibitem{HwangboRAL18}
J.~Hwangbo, J.~Lee, and M.~Hutter, ``Per-contact iteration method for solving
  contact dynamics,'' \emph{IEEE Robot. Autom. Lett.}, vol.~3, no.~2, pp.
  895--902, 2018.

\bibitem{OrtegaSIAM20}
J.~M. Ortega and W.~C. Rheinboldt, \emph{Iterative solution of nonlinear
  equations in several variables}.\hskip 1em plus 0.5em minus 0.4em\relax SIAM,
  2000.

\bibitem{ZhangRSS07}
L.~Zhang, Y.~J. Kim, and D.~Manocha, ``A fast and practical algorithm for
  generalized penetration depth computation,'' in \emph{Robot. Sci. Syst.},
  2007.

\bibitem{KSB}
KS-B-0201, ``Metric coarse screw threads,'' \emph{Korean Standard}, 2016.

\bibitem{KimRobotica17}
M.~Kim and D.~J. Lee, ``Improving transparency of virtual coupling for haptic
  interaction with human force observer,'' \emph{Robotica}, vol.~35, no.~2, pp.
  354--369, 2017.

\bibitem{Spong20}
M.~W. Spong, S.~Hutchinson, and M.~Vidyasagar, \emph{Robot modeling and
  control}.\hskip 1em plus 0.5em minus 0.4em\relax John Wiley \& Sons, 2020.

\bibitem{LeeTRO19}
T.~Lee, P.~M. Wensing, and F.~C. Park, ``Geometric robot dynamic
  identification: A convex programming approach,'' \emph{IEEE Trans. Robot.},
  vol.~36, no.~2, pp. 348--365, 2019.

\bibitem{Nene97TPAMI}
S.~A. Nene and S.~K. Nayar, ``A simple algorithm for nearest neighbor search in
  high dimensions,'' \emph{IEEE Trans. Pattern Anal. Mach. Intell.}, vol.~19,
  no.~9, pp. 989--1003, 1997.

\bibitem{Liu05TASE}
T.~Liu and M.~Y. Wang, ``Computation of three-dimensional rigid-body dynamics
  with multiple unilateral contacts using time-stepping and gauss-seidel
  methods,'' \emph{IEEE Trans. Autom. Sci. Eng.}, vol.~2, no.~1, pp. 19--31,
  2005.

\bibitem{lee2012passive}
D.~J. Lee, M.~Kim, and T.~Qiu, ``Passive haptic rendering and control of
  lagrangian virtual proxy,'' in \emph{IEEE/RSJ Int. Conf. Intell. Robots
  Syst.}, 2012, pp. 64--69.

\bibitem{kim2013toward}
M.~Kim and D.~J. Lee, ``Toward transparent virtual coupling for haptic
  interaction during contact tasks,'' in \emph{World Haptics Conf.}, 2013, pp.
  163--168.

\bibitem{LeeTRO13}
K.~Huang and D.~J. Lee, ``Consensus-based peer-to-peer control architecture for
  multiuser haptic interaction over the internet,'' \emph{IEEE Trans. Robot.},
  vol.~29, no.~2, pp. 417--431, 2013.

\end{thebibliography}
\end{document}